\documentclass[runningheads]{llncs}

\usepackage[mobile]{eccv}

\usepackage{eccvabbrv}

\usepackage{graphicx}
\usepackage{booktabs}

\usepackage[accsupp]{axessibility}  

\usepackage[pagebackref,breaklinks,colorlinks,citecolor=eccvblue]{hyperref}

\usepackage{orcidlink}

\begin{document}

\title{M$^2$Depth: Self-supervised Two-Frame \underline{M}ulti-camera \underline{M}etric Depth Estimation} 

\titlerunning{M$^2$Depth}

\author{Yingshuang Zou \inst{1,2}\textsuperscript{*} \quad Yikang Ding \inst{2}\textsuperscript{*\textdagger} \quad Xi Qiu \inst{2} \quad \\ 
Haoqian Wang \inst{1}\textsuperscript{\textdaggerdbl} \quad Haotian Zhang\inst{2}\textsuperscript{\textdaggerdbl} }

\authorrunning{Y.~Zou et al.}

\institute{$^1$ Tsinghua University $^2$ Megvii Technology}

\maketitle
\let\thefootnote\relax\footnotetext{\textsuperscript{*} Equal Contribution. \textsuperscript{\textdagger}Project Leader. \textsuperscript{\textdaggerdbl}Co-corresponding Authors.}

\begin{abstract}
  This paper presents a novel self-supervised two-frame multi-camera metric depth estimation network, termed M${^2}$Depth, which is designed to predict reliable scale-aware surrounding depth in autonomous driving. Unlike the previous works that use multi-view images from a single time-step or multiple time-step images from a single camera, M${^2}$Depth takes temporally adjacent two-frame images from multiple cameras as inputs and produces high-quality surrounding depth.
  We first construct cost volumes in spatial and temporal domains individually and propose a spatial-temporal fusion module that integrates the spatial-temporal information to yield a strong volume presentation.
  We additionally combine the neural prior from SAM features with internal features to reduce the ambiguity between foreground and background and strengthen the depth edges.
  Extensive experimental results on nuScenes and DDAD benchmarks show M${^2}$Depth achieves state-of-the-art performance. More results can be found in \href{https://heiheishuang.xyz/M2Depth/}{project page}. 
\end{abstract}

\section{Introduction}
\label{sec:intro}


Depth estimation aims to recover the 3D structure of the real world from 2D images, playing a fundamental role in various applications.
In recent years, with the development of autonomous driving, using depth estimation methods to get the 3D representation of the driving scenes shows tremendous attraction, as replacing the expensive depth sensor (\textit{e.g.} Lidar) with vehicle-mounted cameras is cost-effective.
Many previous works~\cite{monodepth2, watson2021manydepth, guizilini2022depthformer, yuan2022new} focus on estimating depth from a single RGB image. Though flexible and concise, such methods suffer from obtaining consistent scale-aware depth (\textit{i.e.} metric depth) among multi-frame and multi-camera when applied in driving scenes.
In order to simultaneously predict the surrounding depth, recent methods~\cite{wei2023surrounddepth, guizilini2022full_surround_monodepth} feed multiple images from $360^o$ vehicle-mounted cameras into 2D encoder-decoder network to capture the spatial information between surround cameras. However, these methods use only one frame and ignore the temporal information, still facing the challenge of predicting consistent metric depth.
Some existing methods~\cite{li2023dynamic_multiframe_depth, wang2023crafting} show that taking temporally adjacent frames as inputs could help get reliable depth under the single camera setting. Nevertheless, few works explore and leverage the spatial-temporal information to strengthen the surrounding depth estimation in driving scenes.

\begin{figure}[tp]
  \centering
   \includegraphics[width=0.9\linewidth]{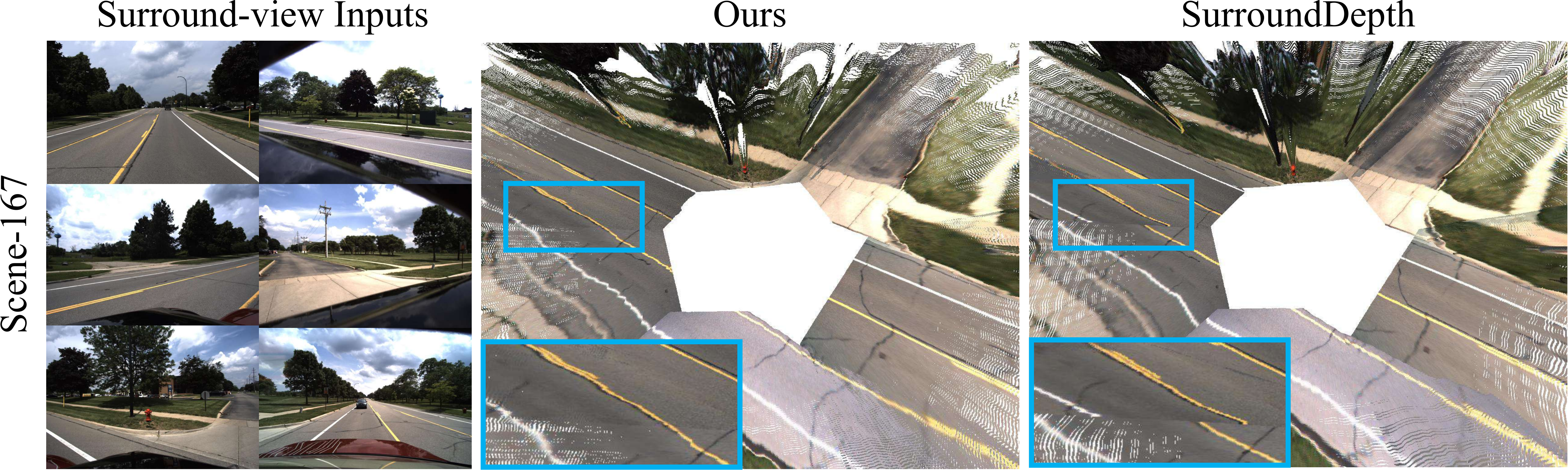}
   \caption{
   Point clouds comparison on  DDAD~\cite{guizilini2020ddad_packnet} dataset. By transforming the predicted depth into point clouds, we show that our method achieves more consistent and accurate estimation compared with SurroundDepth~\cite{wei2023surrounddepth}. The visualized point clouds are fused using the surrounding depth of one frame, and the blue boxes highlight the challenging area spanning multiple cameras.}
   \vspace{-0.5cm}
   \label{fig:compare_points}
\end{figure}

In this paper, we propose a novel self-supervised Two-frame \textit{M}ulti-camera \textit{M}etric depth estimation network, named M$^2$Depth, to predict consistent scale-aware surrounding depth.
The key insight of M$^2$Depth is that we believe combining the spatial-temporal info could boost the surrounding depth estimation, as the spatial info provides an important world scale (from calibrated extrinsic between adjacent cameras) and the temporal info benefits the depth consistency.
As shown in~\cref{fig:compare_points}, M$^2$Depth is able to recover the 3D point clouds with coherence between multiple cameras, while the existing method struggles with keeping multi-camera consistency.
Specifically, M$^2$Depth determines depth by constructing 3D cost volumes within the spatial-temporal domain and applying constraints across multiple cameras, which is different from existing methods~\cite{wei2023surrounddepth,monodepth2,watson2021manydepth,guizilini2022full_surround_monodepth,kim2022vfdepth}.
Following the classical plane-sweeping algorithm~\cite{collins1996plane-sweep}, we construct temporal volumes by utilizing temporal adjacent frames, while the spatial volumes are built by leveraging each view and its overlapped spatial adjacent views.
Building accurate cost volumes faces several challenges. First, getting reliable relative pose and depth annotations is difficult, thus we design a pose estimation branch to predict the relative vehicle pose between two frames and train M$^2$Depth in a self-supervised manner. Second, the depth range in driving scenes is typically large, we consequently design a mono prior branch to estimate coarse depth to narrow down the depth search range.
The initial 3D cost volumes are constructed in the spatial domain and temporal domain separately, which consist of the co-visibility information on the spatially and temporally adjacent views. To jointly use the spatial-temporal clues, we propose a novel spatial-temporal fusion (STF) module, which fuses the initial volumes with visibility-aware weights.
As a result, the fused volumes integrate the space-time correlation between multiple frames and multiple cameras, which will be then decoded to produce the final depth. 

Additionally, we observe that the feature learning of M$^2$Depth is unstable as it actually learns to simultaneously estimate the relative pose, monocular prior, and the multi-camera depth under weak supervision. 
Specifically, we find that the depth estimation method of constructing spatial-temporal volume through pixel matching lacks consistency within instances and discrimination between instances for features. Due to poor features that lack the discrimination between instances, the quality of the depth map decreases.
Inspired by the Segment Anything Model (\textit{a.k.a.} SAM)~\cite{kirillov2023segment_anything}, we propose to inject the strong neural prior from pretrained SAM features into depth estimation to strengthen the feature learning.
The key insight is that SAM is able to capture fine-grained inter-view and intra-view semantic information, which is critical for surrounding depth estimation. We thus design a multi-grained feature fusion (MFF) module to integrate SAM features.
To the best of our knowledge, M$^2$Depth is the first to use the SAM feature in a depth estimation task.


We train and validate M$^2$Depth on two large-scale multi-camera depth estimation benchmarks, \textit{i.e.} DDAD~\cite{guizilini2020ddad_packnet} and nuScenes~\cite{caesar2020nuscenes}, and the extensive experimental results demonstrate M$^2$Depth achieves state-of-the-art performance in multi-camera metric depth estimation task.

In summary, our main contributions are as follows:
\begin{itemize}
    \item We present M$^2$Depth, a novel self-supervised two-frame multi-camera metric depth estimation network, which achieves state-of-the-art performance on multiple surrounding depth estimation benchmarks.
    \item For the first time, we propose to construct spatial-temporal 3D cost volumes and design a spatial-temporal fusion (STF) module for surrounding depth estimation, which strengthens the depth accuracy by fusing the spatial-temporal information. 
    \item We introduce the strong SAM prior into the depth estimation task and propose a multi-grained feature fusion (MFF) module to integrate SAM features with internal features for enhancing the depth quality in detail.
\end{itemize}

\section{Related Works}\label{sec:related_works}

\begin{figure}[htp]
    \centering
    \includegraphics[width=\linewidth]{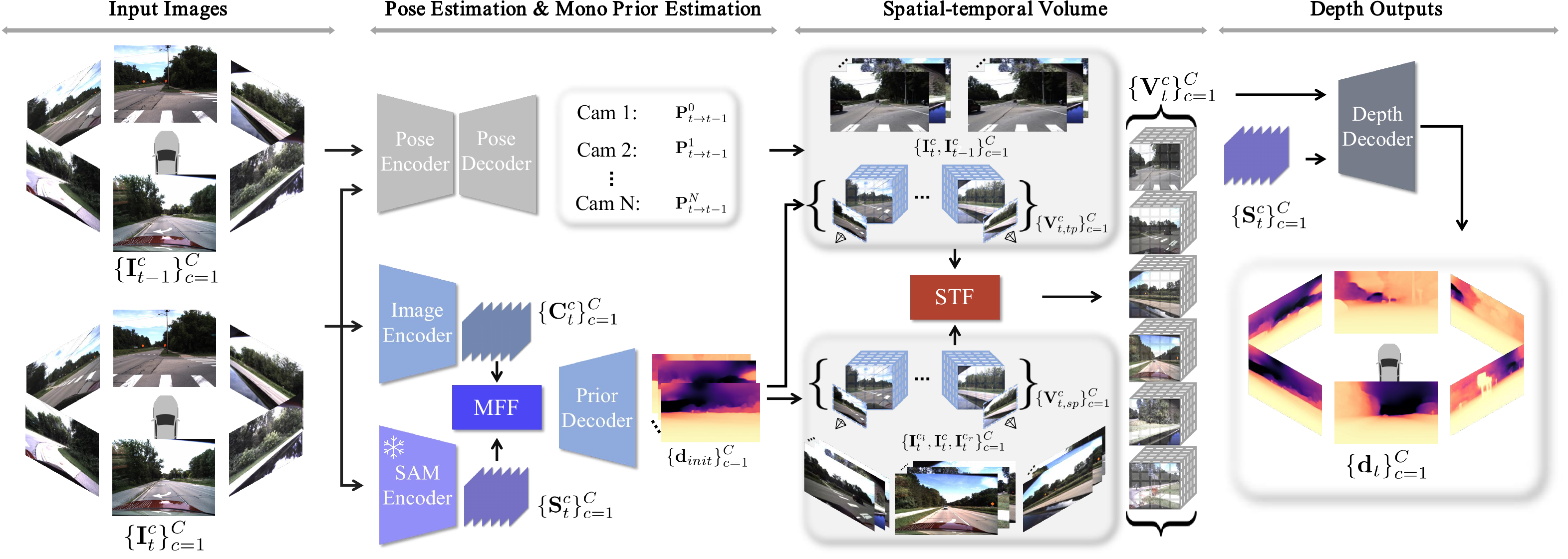}
    \caption{Overview of M$^2$Depth. Given images $\{\mathbf{I}_{t}^c\}_{c=1}^C$ and $\{\mathbf{I}_{t-1}^c\}_{c=1}^C$ from multiple cameras and two frames, M$^2$Depth first estimates the pose of the front camera $\mathbf{P}_{t \to t-1}$, which will be used to infer the poses of all other cameras $\{\mathbf{P}^0_{t \to t-1}\}_{c=1}^C$. 
    In mono prior estimation, the multi-grained feature fusion (MFF) module aggregates the internal features $\{\mathbf{C}_t^c\}_{c=1}^C$ from image encoder and the SAM features $\{\mathbf{S}_t^c\}_{c=1}^C$ from SAM encoder to improve feature expression in multi-grained. The depth prior and constraints across multiple cameras are employed to construct 3D cost volumes $\{\mathbf{V}_t^c\}_{c=1}^C$ within the temporal-spatial domain, which will be then used by the spatial-temporal fusion (STF) module to strengthen the accuracy and consistency of cost volumes. Finally, the depth decoder takes as inputs the $\{\mathbf{V}_t^c\}_{c=1}^C$ and $\{\mathbf{S}_t^c\}_{c=1}^C$ to produce the surrounding depth.}
    \label{fig:pipeline}
    \vspace{-0.5cm}
\end{figure}

\subsection{Multi-frame Depth Estimation}
Unlike the monocular depth estimation (MDE) works~\cite{monodepth2, bhat2021adabins, li2022binsformer, zhang2023lite} that solely use single frame to predict depth, multi-frame depth estimation works~\cite{wimbauer2021monorec, yang2018deep, watson2021manydepth, bae2022magnet, guizilini2022depthformer, bian2019unsupervised} take as inputs the adjacent frames to enhance depth quality, which yields great improvements in practical applications.
MonoRec~\cite{wimbauer2021monorec} constructs a cost volume based on multiple frames from a single camera to estimate depth. It additionally requires a visual odometry system~\cite{yang2018deep} to provide inter-frame pose and sparse depth as supervision signals.
Manydepth~\cite{watson2021manydepth} learns adaptive cost volume from input data and proposes consistency loss for moving objects.
Some methods~\cite{li2023dynamic_multiframe_depth, wang2023crafting, feng2022disentangling} attempt to estimate a depth prior and then encode it into multi-frame cues. 

Despite utilizing multiple frames as input, these methods still struggle to produce reliable surrounding depth when applied in multi-frame multi-camera scenarios. In contrast, our approach imposes spatial constraints via the overlaps among multiple cameras, achieving consistent scale-aware surrounding depth estimation.

\subsection{Multi-camera Depth Estimation}
Intelligent vehicles are typically equipped with multiple surrounding cameras, thus recovering the surrounding 3D environment from mounted cameras turns into a fundamental task in autonomous driving.
FSM~\cite{schonberger2016structure_from_motion} pioneers the expansion of self-supervised monocular depth estimation to surrounding views by using temporal texture constraints as supervision. 
SurroundDepth~\cite{wei2023surrounddepth} employs the cross-view transformer module to perform feature interaction in surrounding views.
EGA-Depth~\cite{shi2023ega} optimizes the computational cost of the attention module, enabling the utilization of higher-resolution feature maps. 
VFDepth~\cite{kim2022vfdepth} constructs a unified volumetric feature representation to estimate the surrounding depth and canonical vehicle pose.

Unlike the aforementioned methods that use a single frame to predict depth, we construct cost volumes using multi-camera and two frames in the spatial-temporal domain. 
It is noteworthy that the concurrent work R3D3~\cite{schmied2023r3d3} uses the SLAM algorithm~\cite{teed2021droid} and a sequence of frames to estimate vehicle pose and optical flow, which will be then utilized to produce depth. Such a manner inevitably needs many frames as inputs and consumes more computation. On the contrary, our method could use only two frames to achieve comparable performance.

\subsection{SAM and Applications}
Segment Anything Model~\cite{kirillov2023segment_anything} (SAM) is an effective, promptable transformer-based model for image segmentation, trained on the SA-1B dataset~\cite{kirillov2023segment_anything}, which comprises over 1 billion masks on 11M licensed and privacy respecting images.
Its exceptional performance in fine-grained semantic segmentation establishes SAM as a prominent player in numerous tasks, such as tracking~\cite{cheng2023segment_and_tracking}, image inpainting~\cite{yu2023inpaint}, video text spotting~\cite{he2023video_text_spotting}, medical image domain~\cite{Zhang2023samed, ma2023MedSAM, bui2023sam3d}.
SAMFeat~\cite{wu2023samfeat} applies SAM to segmentation-independent visual tasks and 
improve local feature description by using feature-level distillation.

To the best of our knowledge, we are the first to apply SAM in the depth estimation task. By leveraging the SAM feature, we extract fine-grained semantic information from and enhance the accuracy of surrounding depth.



\section{Methods}\label{sec:method}

\subsection{Problem Formulation}\label{subsec:problem_form}
This paper focuses on the surrounding depth estimation task in autonomous driving, where the cameras are mounted on ego vehicles and provide $360^{o}$ visual observations. We define that there are $C$ cameras with known intrinsics $\{\mathbf{K}^c\}_{c=1}^C$ and extrinsics $\{\mathbf{T}^c\}_{c=1}^C$, which associate the cameras with the ego vehicle.
Given images of the current frame $\{\mathbf{I}_{t}^c\}_{c=1}^C$ and the previous frame $\{\mathbf{I}_{t-1}^c\}_{c=1}^C$
from surrounding cameras, M$^2$Depth produces the scale-aware surrounding depth $\{\mathbf{d}_t^c\}^C_{c=1}$ at the current timestamp $t$.
It is noteworthy that the ground truth relative pose of the vehicle between two frames is not required, M$^2$Depth is able to predict the relative pose with scale, which is inherently stored in $\{\mathbf{T}^c\}_{c=1}^C$ and the overlap between adjacent views.


\subsection{Network Overview}\label{subsec:net_overview}

The overall architecture of M$^2$Depth is illustrated in \cref{fig:pipeline}.
The input images $\{\mathbf{I}_{t}^c\}_{c=1}^C$ and $\{\mathbf{I}_{t-1}^c\}_{c=1}^C$ are first used to perform pose estimation and mono prior estimation. Specifically, the pose encoder takes the images of the front view ($\mathbf{I}_{t}^0$ and  $\mathbf{I}_{t-1}^0$) as input and learns to predict the 6-Dof relative pose between $\mathbf{I}_{t}^0$ and  $\mathbf{I}_{t-1}^0$.
Unlike previous methods that concatenate surrounding views to directly predict the ego pose~\cite{wei2023surrounddepth} or construct 4D volumes to estimate optical flow and calculate the ego pose~\cite{schmied2023r3d3}, our approach simplifies the ego pose estimation problem by estimating the front camera's pose $\mathbf{P}^0_{t \to t-1}$, thus the ego pose $\mathbf{P}_{t \to t-1}$ can be derived by:
\begin{equation}
    \mathbf{P}_{t \to t-1} = {(\mathbf{T}^0})^{-1} \mathbf{P}^0_{t \to t-1} \mathbf{T}^0,
    \label{eq:pose_cal}
\end{equation}
where $({\mathbf{T}^0})^{-1}$ indicates the inverse matrix of $\mathbf{T}^0$, and $\mathbf{T}^0$ is the extrinsic matrix between the front camera and ego vehicle.

In monocular prior estimation (\cref{subsec:sam_mono}), the $\mathbf{I}_{t}^0$ is fed into a trainable image encoder and a frozen SAM encoder, where the output features are then fused in multi-grained feature fusion (MFF) module. MFF aims to integrate the fine-grained semantic information in SAM features with the depth cues in internal features, which helps M$^2$Depth understand the 3D environment. As a result, the prior decoder takes fused features as inputs and produces the mono prior depth $\{{\mathbf{d}_{t}^c}_{prior}\}^{C}_{c=1}$, which plays the role of depth guidance in volume construction.


After obtaining the relative vehicle pose $\mathbf{P}_{t \to t-1}$ and the mono prior $\{{\mathbf{d}_{t}^c}_{prior}\}^{C}_{c=1}$, M$^2$Depth constructs the spatial-temporal cost volumes (\cref{subsec:ray_attention}).
Specifically, we use the plane-sweeping algorithm~\cite{collins1996plane-sweep} to construct the initial cost volume. 
We first employ a feature pyramid network~\cite{lin2017fpn} to extract matching features $\{ \mathbf{F}_{t}^c\}_{c=1}^C$ from ${\{\mathbf{I}_{t}^c\}}_{c=1}^C$.
As for the temporal domain, we warp the feature $\mathbf{F}_{t-1}^c$, which is decoded from $\mathbf{I}_{t-1}^c$, to $\mathbf{F}_{t}^c$ according to the sampled depth values $d_{t}^c$ to get the temporal volume $\mathbf{V}_{t,tp}^c$, where $d_{t}^c$ is appointed based on ${\mathbf{d}_{t}^c}_{prior}$. Similarly, the spatial volumes $\mathbf{V}_{t,sp}^c$ are constructed using spatial adjacent views $\mathbf{F}_{t}^{c_l}$ and $\mathbf{F}_{t}^{c_r}$, where the $c_l$ and $c_r$ represent the adjacent left and right camera of the reference camera.
The initial volumes $\{\mathbf{V}_{t,tp}^c\}_{c=1}^{C}$ and $\{\mathbf{V}_{t,sp}^c\}_{c=1}^{C}$ will be fed into the spatial-temporal fusion (STF) module, which fuses the spatial-temporal information and yields the final volumes $\{{\mathbf{V}_t^c\}_{c=1}^{C}}$.
Subsequently, we decode the $\{{\mathbf{V}_t^c\}_{c=1}^{C}}$ into depth probability distribution volumes, and produce the estimated depth $\{{\mathbf{d}_{t}^c}\}^{C}_{c=1}$ by calculating the depth expectation. 

\subsection{Mono Prior Estimation}\label{subsec:sam_mono}
Following the plane sweep paradigm, the two-frame depth estimation problem can be transformed into a feature matching task~\cite{ding2022transmvsnet,gu2020casmvsnet}, where the depth samples would significantly affect final depth quality.  
Unfortunately, the total depth range in driving scenarios is typically large. As a result, it requires a lot of depth samples to predict precise depth, which would cost tremendous computation in multi-camera settings. To handle this problem, we use a monocular prior estimation branch to produce coarse guidance for cost volume construction.

Specifically, given surrounding views in the current timestamp ${\{\mathbf{I}_{t}^c\}}_{c=1}^C$, we first use a CNN encoder and a SAM encoder~\cite{kirillov2023segment_anything} to extract internal features $\{\mathbf{C}_t^c\}_{c=1}^C$ and SAM priors $\{\mathbf{S}_t^c\}_{c=1}^C$, respectively.
Then we use the Multi-grained Feature Fusion (MFF) module to fuse $\{\mathbf{C}_t^c\}_{c=1}^C$ and $\{\mathbf{S}_t^c\}_{c=1}^C$, and finally use a CNN decoder to generate mono depth prior $\{{\mathbf{d}_{t}^c}_{prior}\}^{C}_{c=1}$.

\subsubsection{Multi-grained Feature Fusion}
\begin{figure}[t]
  \centering
   \includegraphics[width=0.78\linewidth]{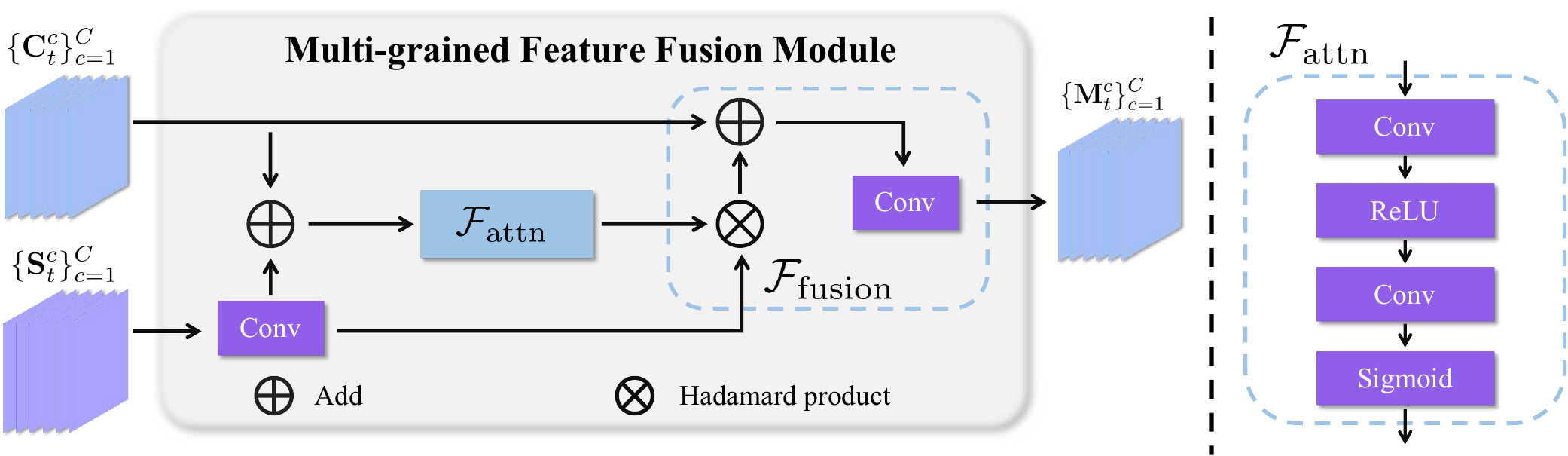}
   \caption{
   Details of the multi-grained feature fusion (MFF) module. MFF takes the internal features $\{\mathbf{C}_t^c\}_{c=1}^C$ and the SAM features $\{\mathbf{S}_t^c\}_{c=1}^C$ as inputs, and utilizes a $\mathcal{F}_{\mathrm{attn}}$ block to yield the weight map, which fetches the complementary info between $\{\mathbf{C}_t^c\}_{c=1}^C$ and $\{\mathbf{S}_t^c\}_{c=1}^C$, and will be used in $\mathcal{F}_{\mathrm{fusion}}$ block to produce the fused feature $\{\mathbf{M}_t^c\}_{c=1}^C$.
   }
   \label{fig:mgfusion_moudle}
\end{figure}

The detailed structure of the MFF module is illustrated in~\cref{fig:mgfusion_moudle}, which is the key part of depth prior estimation.
Given an internal feature $\mathbf{C}_t^c \in \mathbb{R}^{H\times W\times C}$ and a SAM feature $\mathbf{S}_t^c \in\mathbb{R}^{H\times W\times C'}$ of a certain camera, we first use convolution layers to align the dimension of $\mathbf{S}_t^c$ with $\mathbf{C}_t^c$, which will be then combined and fed into the $\mathcal{F}_{\mathrm{attn}}$ to yield attention weights $\mathbf{W}_t^c$:
\begin{equation}
    \mathbf{W}_t^c = \mathcal{F}_{\mathrm{attn}}(\mathbf{C}_t^c, \mathbf{S}_t^c) = \sigma (f^{3 \times 3}(\delta(f^{3 \times 3}(\mathbf{C}_t^c + \mathbf{S}_t^c)))), 
    \label{eq:attention_weight}
\end{equation}
where the $\sigma$ refers to the $sigmoid$ function, $\delta$ refers to the $ReLU$~\cite{nair2010relu} function, and $f^{3 \times 3}$ denotes the convolution layer with kernel size of $3 \times 3$.
Intuitively, the $\mathbf{W}_t^c$ fetches the complementary info between $\mathbf{S}_t^c$ and $\mathbf{C}_t^c$, and performs as a feature guidance in feature fusion: 
\begin{equation}
    \mathbf{M}_t^c = \mathcal{F}_{\mathrm{fusion}}(\mathbf{C}_t^c, \mathbf{S}_t^c, \mathbf{W}_t^c) = f^{3 \times 3}(\mathbf{C} + \mathbf{W}_t^c \odot \mathbf{S}_t^c), 
    \label{eq:mono_feature}
\end{equation}
where $\mathbf{M}_t^c$ represents the fused features and $\odot$ denotes the $Hadamard$ product.
As introducing SAM features into mono prior estimation helps get better performance, we further utilize SAM features in the depth decoding phase in a similar manner to mono prior estimation, more details can be found in supplementary materials, and experimental evaluation is conducted in~\cref{sec:exp}.

\subsection{Spatial-temporal Cost Volume}\label{subsec:ray_attention}
Taking $\mathbf{I}_t^c \in \mathbb{R}^{H \times W \times 3}$ of a certain camera for example, we denote its spatial adjacent views as $\mathbf{I}_{t}^{c_l}, \mathbf{I}_{t}^{c_r}$ and its last frame as $\mathbf{I}_{t-1}^{c}$.
In this section, we use the estimated relative pose $\mathbf{P}_{t \to t-1}$ pose and the known camera extrinsics $\{\mathbf{T}^c\}_{c=1}^C$ to construct the spatial-temporal cost volume $\{\mathbf{V}_{t}^c\}_{c=1}^C$.

\subsubsection{Initial Volume Construction}\label{subsubsec:initial_volume}

Taking the aforementioned images as input, we first employ a Feature Pyramid Network (FPN)~\cite{lin2017fpn} to extract image features $\mathbf{F}_{t}^{c} \in \mathbb{R}^{\frac{H}{4} \times \frac{W}{4} \times 3}$.
As for spatial cost volume, we warp $\mathbf{F}_{t}^{c_l}$ and $\mathbf{F}_{t}^{c_r}$ into the current camera's frustum according to the camera intrinsics, extrinsics, and the depth samples.
The warping operation between a pixel $\mathbf{p}$ in reference view $c$ and its corresponding pixel $\hat{\mathbf{p}}$ in adjacent views $c' \in \{c_l, c_r\}$ under depth sample $d$ is defined as:
\begin{equation}
    \hat{\mathbf{p}} = \mathbf{K}^{c'} \cdot [\mathbf{R}^{c \to c'} \cdot ((\mathbf{K}^{c})^{-1} \cdot \mathbf{p} \cdot d) + \mathbf{t}^{c \to c'}],
    \label{eq:warp_spatial}
\end{equation}
where the transformation matrix $\mathbf{P}^{c \to c'} = [\mathbf{R} | \mathbf{t}]^{c \to c'}$ can be written as:
\begin{equation}
    \mathbf{P}^{c \to c'} = (\mathbf{T}^{c'})^{-1} \cdot \mathbf{T}^c,
    \label{eq:pose_spatial}
\end{equation}
Afterward, we combine the warped features to form the initial spatial volumes $\{\mathbf{V}_{t,sp}^c\}_{c=1}^C$.
The temporal volume $\{\mathbf{V}_{t,tp}^c\}_{c=1}^C$ can be constructed similarly.
Specifically, the warping operation in the temporal domain is defined as:
\begin{equation}
    \hat{\mathbf{p}} = \mathbf{K}^{c} \cdot [\mathbf{R}_{t \to t'}^{c} \cdot ((\mathbf{K}^{c})^{-1} \cdot \mathbf{p} \cdot d) + \mathbf{t}_{t \to t'}^{c}],
    \label{eq:warp_temporal}
\end{equation}
where the $\hat{\mathbf{p}}$ denotes the corresponding pixel in the previous frame, and the $\mathbf{P}_{t \to t'}^c = [\mathbf{R}^{c} | \mathbf{t}^{c}]_{t \to t'}$ is obtained according to~\cref{eq:pose_cal}:
\begin{equation}
    \mathbf{P}_{t \to t'}^c = (\mathbf{T}^{c})^{-1} \cdot \mathbf{P}_{t \to t'} \cdot \mathbf{T}^c.
    \label{eq:pose_temporal}
\end{equation}
During the construction of the initial volumes, we employ $\{\mathbf{d}_{prior}\}^C_{c=1}$ as the depth guidance and appoint the depth samples in an adaptive range, more details of the depth samples can be found in supplementary materials.



\begin{figure}[t]
    \centering
    \includegraphics[width=0.9\linewidth]{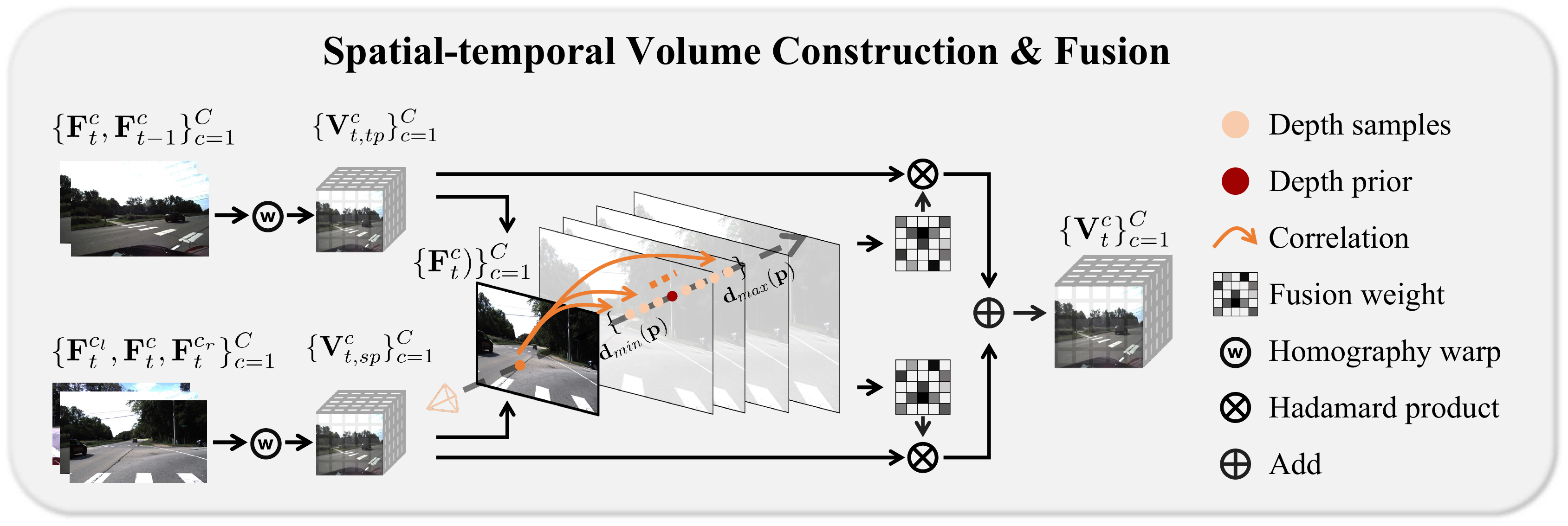}
    \caption{Overview of the volume construction and STF module. Given the reference image feature $\mathbf{F}_{t}^{c}$ and its temporal-spatial adjacent features, we first warp the adjacent features to reference view to form the initial volumes $\mathbf{V}_{t,sp}^{c}$ and $\mathbf{V}_{t,tp}^{c}$ in spatial domain and temporal domain respectively. After that, STF fuses the initial volumes by computing the correlation between $\mathbf{F}_{t}^{c}$ and $\mathbf{V}_{t,sp}^{c}$, $\mathbf{V}_{t,tp}^{c}$ and produces the weight maps $\mathbf{W}_{t,sp}^{c}$, $\mathbf{W}_{t,tp}^{c}$, which will be used as fusion weights to guide the volume fusion.}
    \label{fig:STF}
\end{figure}
\subsubsection{Spatial-temporal Volume Fusion}\label{subsubsec:ray_attention}
By constructing the initial cost volumes, we approach depth estimation as a feature-matching task, where we try to find the best matching feature among the sampled volume features. To strengthen the matching quality using the spatial-temporal information, we propose a spatial-temporal fusion module, which is depicted in~\cref{fig:STF}.


Given the reference feature $\mathbf{F}_{t}^{c}$ and the initial volumes $\mathbf{V}_{t,sp}^{c}$ and $\mathbf{V}_{t,tp}^{c}$, STF produces the fusion weight maps by computing the group-wise correlation~\cite{guo2019group}. Let $\mathbf{p}$ denote the pixel in the reference feature, to compute the correlation between $\mathbf{F}_{t}^{c}(\mathbf{p})$ and $\mathbf{V}_{t,sp}^{c}(\mathbf{p})$, we first divide $C$ feature channels evenly into $G$ groups, $\mathbf{F}_{t}^{c}(\mathbf{p})^g$ and $\mathbf{V}_{t,sp}^{c}(\mathbf{p})^g$, thus the $g$-th group correlation $\mathbf{Cr}_{t,sp}^{c}(\mathbf{p})^g$ can be computed as:
\begin{equation}
    \mathbf{Cr}_{t,sp}^{c}(\mathbf{p})^g = \frac{G}{C}<\mathbf{F}_{t}^{c}(\mathbf{p})^g, \mathbf{V}_{t,sp}^{c}(\mathbf{p})^g>,
\end{equation}
where $<a,b>$ indicates the inner product of $a$ and $b$, and the group correlation $\mathbf{Cr}_{t,tp}^{c}(\mathbf{p})^g$ between $\mathbf{F}_{t}^{c}(\mathbf{p})^g$ and $\mathbf{V}_{t,tp}^{c}(\mathbf{p})^g$ can be obtained with the same manner.
The maximum correlation along group dimension will serve as the final fuse weight $\mathbf{W}_{t,sp}^{c}$ and $\mathbf{W}_{t,tp}^{c}$, which will be used to fuse the final spatial-temporal volume $\mathbf{V}_{t}^{c}$:
\begin{equation}
    \mathbf{V}_{t}^{c} = \mathbf{W}_{t,sp}^{c}\mathbf{V}_{t,sp}^{c} + \mathbf{W}_{t,tp}^{c}\mathbf{V}_{t,tp}^{c}.
\end{equation}

\subsubsection{Depth Prediction}\label{subsubsec:depth_decoder}

The constructed $\{\mathbf{V}_{t}^{c}\}_{c=1}^C$ are finally input to a depth encoder to produce the final depth $\{\mathbf{d}_{t}^{c}\}_{c=1}^C$, which also takes the SAM features $\{\mathbf{S}_t^{c}\}_{c=1}^{C}$ as context feature.
The key operation is that the depth decoder first transforms the $\{\mathbf{V}_{t}^{c}\}_{c=1}^C$ into probability volumes $\{\mathbf{P}_{t}^{c}\}_{c=1}^C$, where the $\mathbf{P}_{t}^{c}(\mathbf{p})$ represents the probability distribution among sampled depth of each pixel $\mathbf{p}$.
The final depth can be obtained by:
\begin{equation}
    \mathbf{d}_{t}^{c}(\mathbf{p}) = \sum_{i=1}^{D} d_i \cdot \mathbf{P}_{t}^{c}(\mathbf{p}, i),
\end{equation}
where the $d_i$ indicates the $i$-th sampled depth and the $\mathbf{P}_{t}^{c}(\mathbf{p}, i)$ is the probability of $\mathbf{p}$ at $i$-th depth.
Please refer to the supplementary materials for more details about the architecture of the depth decoder.


\subsection{Loss Function}\label{subsec:loss_function}
\paragraph{Photometric Loss}
Following the common practice in self-supervised monocular depth estimation works~\cite{monodepth2, wei2023surrounddepth, shi2023ega}, we optimize M$^2$Depth by using the per-pixel photometric error $\mathcal{L}_{\mathrm{photo}}$ as:
\begin{equation}
    \mathcal{L}_{\mathrm{photo}} = \frac{\alpha}{2}(1 - \mathrm{SSIM}(\mathbf{I}_a, \mathbf{I}_b)) + (1 - \alpha)||\mathbf{I}_a - \mathbf{I}_b||_{1},
    \label{eq:photometric_loss}
\end{equation}
where $\mathrm{SSIM}$ is the structural similarity between two images~\cite{wang2004image_ssim}, $\mathbf{I}_a$ and $\mathbf{I}_b$ indicate the ground truth image and the reconstructed image respectively.
It is noteworthy that M$^2$Depth uses $\mathcal{L}_{\mathrm{photo}}$ in both spatial domain and temporal domain, where the spatial photometric error additionally provides the important scale information.

\paragraph{Depth Smoothness Loss}
As in previous works~\cite{monodepth2, wei2023surrounddepth, godard2017loss_smoothness}, we use the edge-aware smoothness loss~\cite{godard2017loss_smoothness} to prevent estimated depth from shrinking:
\begin{equation}
    \mathcal{L}_{\mathrm{smooth}} = |\partial_x \mathbf{d}|e^{-|\partial_x \mathbf{I}|} + |\partial_y \mathbf{d}|e^{-|\partial_y {\mathbf{I}}|}.
    \label{eq:smoothness_loss}
\end{equation}


\paragraph{Depth Edge Loss}
Inspired by~\cite{talker2022mind_the_edge}, we employ edge information derived from images to enhance the quality of depth edges. Given an RGB image $\mathbf{I}$ and its depth map $\mathbf{d}$, we utilize a pre-trained edge detection model~\cite{he2019bcdn} to extract the edge map $\mathbf{E}_{img}$ from $\mathbf{I}$. Subsequently, the edge map $\mathbf{E}_{depth}$ of $\mathbf{d}$ can be calculated by depth gradient. The depth edge loss is defined as:
\begin{equation}
    \mathcal{L}_{\mathrm{edge}} = \mathrm{FL}(\mathbf{E}_{img}, \mathbf{E}_{depth})
    \label{eq:edge_loss}
\end{equation}
where $\mathrm{FL}$ denote the focal loss~\cite{lin2017focal_loss}. 


\paragraph{SfM Loss}
Although the $\mathcal{L}_{\mathrm{photo}}$ in spatial domains could constrain the scale of estimated depth and pose, it relies on good initialization.
Following the previous works~\cite{wei2023surrounddepth, guizilini2022full_surround_monodepth}, we use SfM to generate sparse depth between spatially adjacent views to endow the network with an initial rudimentary estimation of scale-aware depth.
More details of SfM loss $\mathcal{L}_{\mathrm{sfm}}$ can be found in supplementary materials.

\paragraph{Total Loss}
The overall loss function can be written as:
\begin{equation}
    \mathcal{L} = \lambda_1 \mathcal{L}_{\mathrm{photo}} + \lambda_2 \mathcal{L}_{\mathrm{smooth}} + \lambda_3 \mathcal{L}_{\mathrm{edge}} + \lambda_4 \mathcal{L}_{\mathrm{sfm}},
    \label{eq:loss_all}
\end{equation}
where the $\lambda_1$, $\lambda_2$, $\lambda_3$, $\lambda_4$ are weights of different losses, we set $\lambda_1 = 1.0$, $\lambda_2 = 1.0e$-3, $\lambda_3 = 1.0e$-2, $\lambda_4 = 1.0e$-2, and the $\lambda_4$ is set to $0$ after initialization.

\section{Experiments}
\label{sec:exp}

\subsection{Implementation Details} 
We implement M$^2$Depth using PyTorch and train the model using Adam as optimizer with a learning rate set to $10^{-4}$, $\beta_1 = 0.9$ and $\beta_2 = 0.999$.
For the pose estimation, we employ a ResNet-34~\cite{he2016resent} model to predict the axis-angle and translation of the front camera.
For the depth prior estimation branch, we employ the ResNet-34~\cite{he2016resent} model to predict internal features and use the frozen SAM encoder provided by MobileSAM~\cite{zhang2023mobile_sam} for saving memory.
All of our experiments are conducted using 8 NVIDIA V100 GPUs.

\subsection{Dataset}
We train and evaluate M$^2$Depth on two public datasets including DDAD~\cite{guizilini2020ddad_packnet} and nuScenes~\cite{caesar2020nuscenes}.

The dense depth for autonomous driving (DDAD) dataset~\cite{guizilini2020ddad_packnet} is an autonomous driving benchmark that consists of 150 training and 50 validation scenes in complex and diverse urban environments.
Following the previous work~\cite{wei2023surrounddepth, schmied2023r3d3}, we downsample the images from their initial resolution of $1216 \times 1936$ to $384 \times 640$ and evaluate depth up to 200m averaged across all cameras.

The nuScenes dataset~\cite{caesar2020nuscenes} comprises 700 training, 150 validation, and 150 testing urban scenes.
Following the previous work~\cite{wei2023surrounddepth}, we downsample the images from the initial resolution of $900 \times 1600$ to $352 \times 640$ and evaluate depth up to 80m averaged across all cameras.

\subsection{Experimental Results}
\begin{table*}[!t]
\centering
\caption{Quantitative results on DDAD dataset~\cite{guizilini2020ddad_packnet} (evaluate depth up to 200m) and nuScenes dataset~\cite{caesar2020nuscenes} (evaluate depth up to 80m). We present the mean accuracy across all views using the metrics from ~\cite{eigen2014metric}.
The $Frame$ stands for the number of frames in the training\textbackslash{}testing phase.
FSM* indicates the implementation from~\cite{kim2022vfdepth}.
R3D3* denotes the results using the official code and the same frame setting with us. (\textbf{Bold} figures indicate the best and \underline{underlined} figures indicate the second best) }
\vspace{-0.1cm}
\resizebox{\linewidth}{!}{
\begin{tabular}{lccccccccc}
\toprule
\textbf{Method} & Dataset & Frame & Abs. Rel. $\downarrow$ & Sq. Rel. $\downarrow$ & RMSE $\downarrow$ & RMSE log $\downarrow$ & $\delta < 1.25$ $\uparrow$ & $\delta < 1.25^2$ $\uparrow$ & $\delta < 1.25^3$ $\uparrow$ \\
\midrule
R3D3~\cite{schmied2023r3d3}          & \multirow{2}{*}[-2em]{\rotatebox[origin=c]{90}{DDAD~\cite{guizilini2020ddad_packnet}}} & 6\textbackslash{}5 & 0.169    & 3.041   & 11.372 & -        & 0.809 & -     & -     \\
\midrule
FSM~\cite{guizilini2022full_surround_monodepth}           & & 3\textbackslash{}1 & \underline{0.201}    & -       & -      & -        & -     & -     & -     \\
FSM*~\cite{guizilini2022full_surround_monodepth}          & & 3\textbackslash{}1 & 0.228    & 4.409   & 13.433 & 0.342    & 0.687 & 0.870 & 0.932 \\
VFDepth~\cite{kim2022vfdepth}       & & 3\textbackslash{}1 & 0.218    & 3.660   & 13.327 & 0.339    & 0.674 & 0.862 & 0.932 \\
SurroundDepth~\cite{wei2023surrounddepth} & & 3\textbackslash{}1 & 0.208    & \underline{3.371}   & \underline{12.977} & \underline{0.330}     & \underline{0.693} & \underline{0.871} & \underline{0.934} \\
R3D3*~\cite{schmied2023r3d3}         & & 3\textbackslash{}2 & 0.311    & 5.473   & 14.094 & 0.385    & 0.604 & 0.814 & 0.903 \\
Ours         & & 3\textbackslash{}2 & \textbf{0.183}    & \textbf{2.920}    & \textbf{11.963} & \textbf{0.299}    & \textbf{0.756} & \textbf{0.897} & \textbf{0.947} \\
\midrule
R3D3~\cite{schmied2023r3d3}          & \multirow{2}{*}[-2em]{\rotatebox[origin=c]{90}{nuScenes~\cite{caesar2020nuscenes}}} & 6\textbackslash{}5 & 0.253    & 4.759   & 7.150  & -        & 0.729 & -     & -     \\
\midrule
FSM~\cite{guizilini2022full_surround_monodepth}           & & 3\textbackslash{}1 & \underline{0.297}    & -       & -      & -        & -     & -     & -     \\
FSM*~\cite{guizilini2022full_surround_monodepth}          & & 3\textbackslash{}1 & 0.319    & 7.534   & 7.860  & 0.362    & \underline{0.716} & \underline{0.874} & \underline{0.931} \\
VFDepth~\cite{kim2022vfdepth}       & & 3\textbackslash{}1 & 0.289    & 5.718   & 7.551  & \underline{0.348}    & 0.709 & \textbf{0.876} & \textbf{0.932} \\
SurroundDepth~\cite{wei2023surrounddepth} & & 3\textbackslash{}1 & 0.280    & \textbf{4.401}   & \underline{7.467}  & 0.364    & 0.661 & 0.844 & 0.917 \\
R3D3*~\cite{schmied2023r3d3}         & & 3\textbackslash{}2 & 0.498    &  5.489  & 11.740  & 0.746    & 0.155  & 0.375 & 0.613   \\
Ours          & & 3\textbackslash{}2 & \textbf{0.259}    & \underline{4.599}   & \textbf{6.898}  & \textbf{0.332}    & \textbf{0.734} & 0.871 & 0.928 \\
\bottomrule
\end{tabular}}
\label{tab:all_results}
\end{table*}
This paper focuses on scale-aware surrounding depth estimation task, thus we only report the scale-aware results and mainly compare M$^2$Depth with the recent self-supervised surrounding depth estimation methods, including FSM~\cite{guizilini2022full_surround_monodepth}, SurroundDepth~\cite{wei2023surrounddepth}, VFDepth~\cite{kim2022vfdepth} and R3D3~\cite{schmied2023r3d3}, without comparing with the numerous MDE methods~\cite{monodepth2,watson2021manydepth,bhat2021adabins,wang2023sqldepth}.

\begin{table}[th]
    \centering
    \caption{Per-camera evaluation on DDAD dataset~\cite{guizilini2020ddad_packnet}. SD is the abbreviation of SurroundDepth~\cite{wei2023surrounddepth}. R3D3* indicates the results using its official code and the same frame setting with us. Our method achieves superior overall performance across multiple cameras to existing works. According to the memory and computation analysis, M$^2$Depth achieves a good balance between overall performance and computational efficiency.}
    \vspace{-0.2cm}
    \resizebox{\linewidth}{!}{
    \begin{tabular}{l|ccccccc|ccc}
    \toprule
           & \multicolumn{6}{c}{~~~~~~Abs.Rel. $\downarrow$}  & & \multicolumn{3}{c}{Memory \& Computation}    \\
    Method & Front    & F.Left & F.Right & B.Left & B.Right & Back   & Avg.   & Memory(MB)          & Flops(G)    & Time(s)   \\
    \midrule
    FSM~\cite{guizilini2022full_surround_monodepth}    & \textbf{0.130} & \underline{0.201} & \underline{0.224} & 0.229 & 0.240 & \underline{0.186} & \underline{0.201} & -   & -   & -  \\
    SD~\cite{wei2023surrounddepth}     & 0.152  & 0.207 & 0.230 & \underline{0.220} & \underline{0.239} & 0.200 & 0.208 & \textbf{3042} & \textbf{237.106} & \textbf{0.215}    \\
    R3D3*~\cite{schmied2023r3d3}  & 0.234  & 0.284 & 0.355 & 0.347 & 0.392 & 0.255 & 0.311 & 8371 & 2621.738 & 0.378 \\
    Ours   & \underline{0.146} & \textbf{0.182} & \textbf{0.200} & \textbf{0.198} & \textbf{0.203} & \textbf{0.169} & \textbf{0.183} & \underline{5546} & \underline{866.019} & \underline{0.295}    \\
    \bottomrule
    \end{tabular}}
    \vspace{-0.2cm}
    \label{tab:each_camera}
\end{table}

\paragraph{Results on DDAD}
Following the common practice, we report the quantitative results of $Abs. Rel.$, $Sq. Rel.$, $RMSE$, $RMSE~log$ and $\delta$ as shown in~\cref{tab:all_results}. The specific definition of the evaluation metrics can be found in supplementary materials.
Previous works~\cite{wei2023surrounddepth,kim2022vfdepth} typically use three frames [$t-1,t,t+1$] for training, where the $t+1$ frame is only used in computing loss, we follow this paradigm to train M$^2$Depth.
It is noteworthy that the R3D3~\cite{schmied2023r3d3} takes sequence frames as input, we test its results with 2 frames using their official code for a fair comparison.

\begin{figure*}[!t]
\setlength\tabcolsep{0.5 pt}
\centering
\scalebox{0.8}{
\begin{tabular}{lcccccc}

& Front & F.Left & B.Left & Back & B.Right & F.Right \\
\rotatebox[origin=c]{90}{Input} &
\begin{tabular}{l}\includegraphics[width=0.198\linewidth]{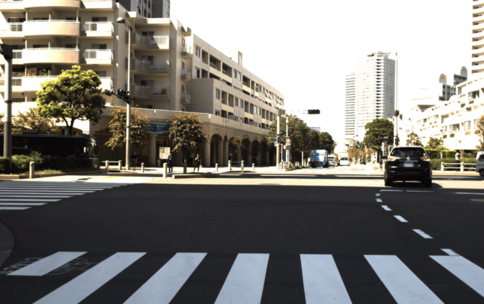}\end{tabular} &
\begin{tabular}{l}\includegraphics[width=0.198\linewidth]{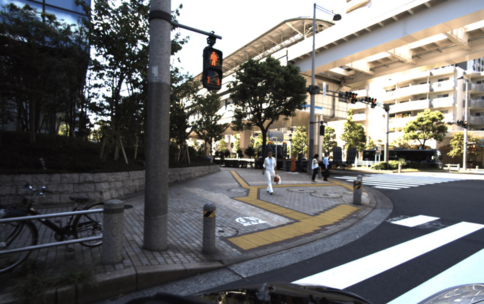}\end{tabular} &
\begin{tabular}{l}\includegraphics[width=0.198\linewidth]{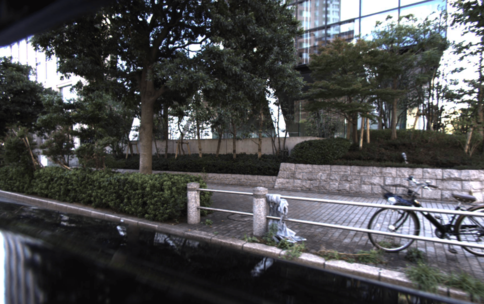}\end{tabular} &
\begin{tabular}{l}\includegraphics[width=0.198\linewidth]{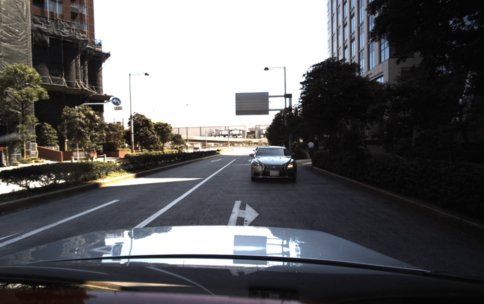}\end{tabular} &
\begin{tabular}{l}\includegraphics[width=0.198\linewidth]{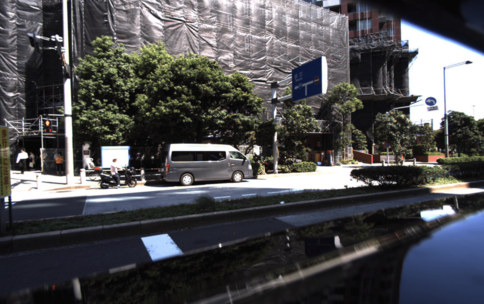}\end{tabular} &
\begin{tabular}{l}\includegraphics[width=0.198\linewidth]{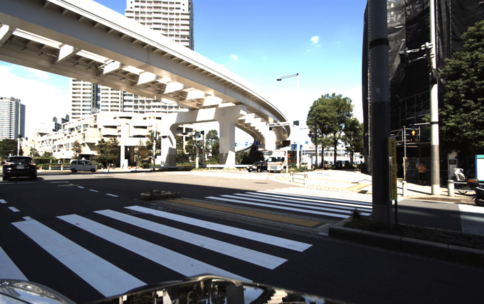}\end{tabular} \\


\multirow{2}{*}[1.2em]{\rotatebox[origin=c]{90}{SurroundDepth~\cite{wei2023surrounddepth}}} &
\begin{tabular}{l}\includegraphics[width=0.198\linewidth]{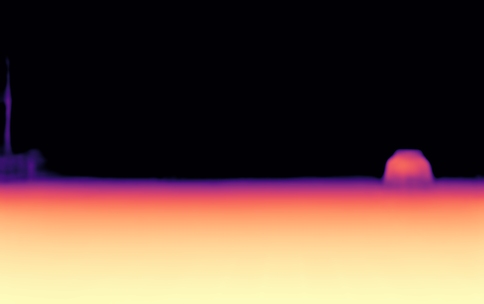}\end{tabular} &
\begin{tabular}{l}\includegraphics[width=0.198\linewidth]{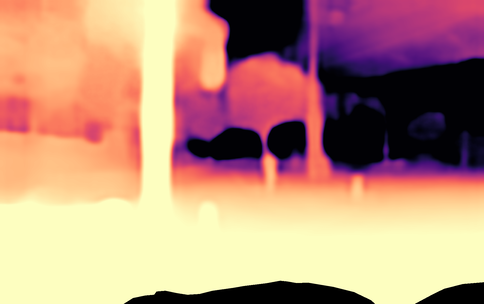}\end{tabular} &
\begin{tabular}{l}\includegraphics[width=0.198\linewidth]{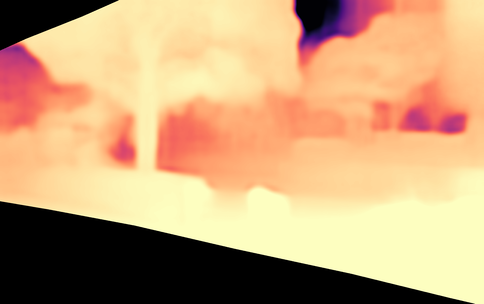}\end{tabular} &
\begin{tabular}{l}\includegraphics[width=0.198\linewidth]{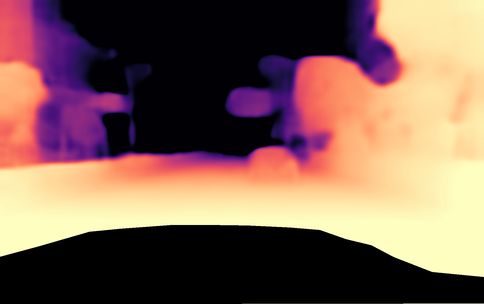}\end{tabular} &
\begin{tabular}{l}\includegraphics[width=0.198\linewidth]{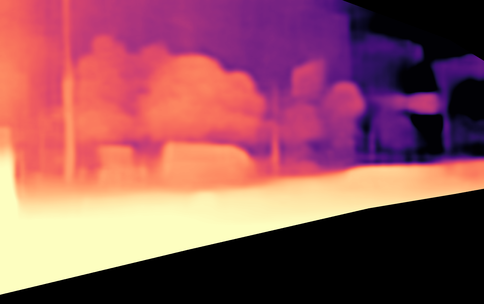}\end{tabular} &
\begin{tabular}{l}\includegraphics[width=0.198\linewidth]{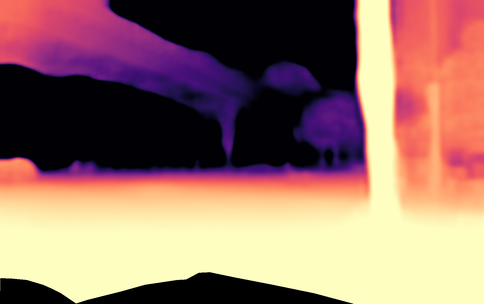}\end{tabular} \\
&
\begin{tabular}{l}\includegraphics[width=0.198\linewidth]{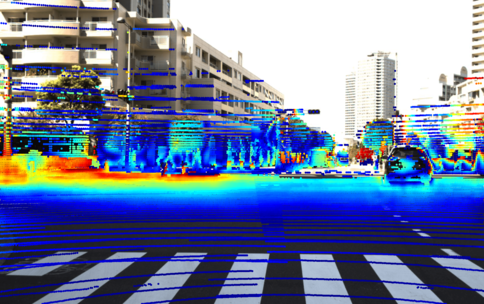}\end{tabular} &
\begin{tabular}{l}\includegraphics[width=0.198\linewidth]{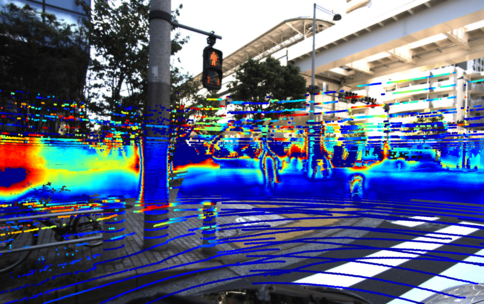}\end{tabular} &
\begin{tabular}{l}\includegraphics[width=0.198\linewidth]{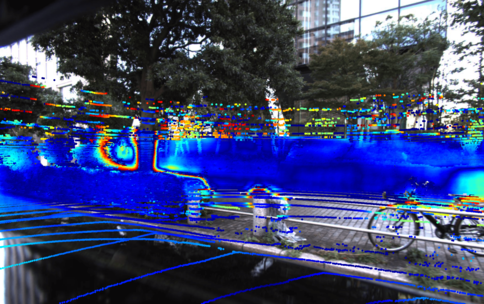}\end{tabular} &
\begin{tabular}{l}\includegraphics[width=0.198\linewidth]{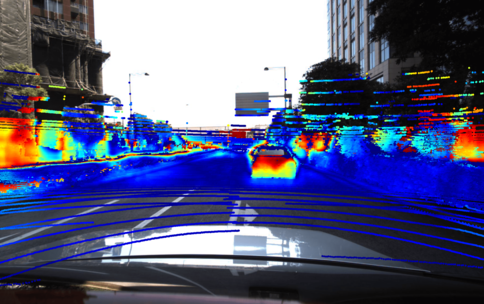}\end{tabular} &
\begin{tabular}{l}\includegraphics[width=0.198\linewidth]{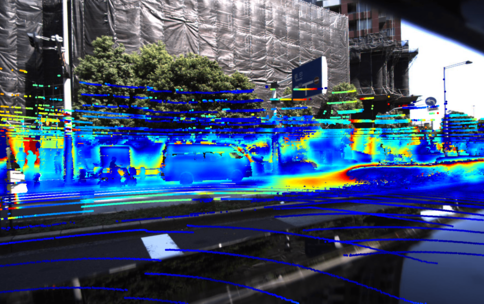}\end{tabular} &
\begin{tabular}{l}\includegraphics[width=0.198\linewidth]{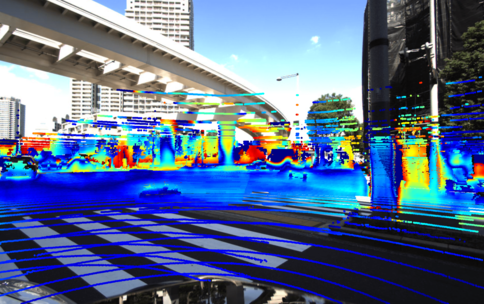}\end{tabular} \\


\multirow{2}{*}[-1.5em]{\rotatebox[origin=c]{90}{Ours}} &
\begin{tabular}{l}\includegraphics[width=0.198\linewidth]{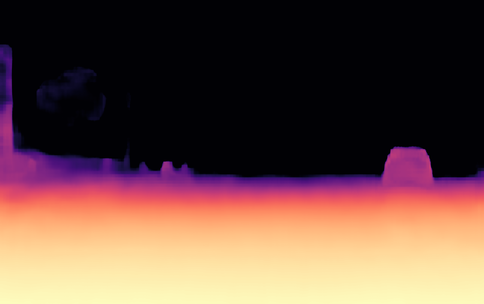}\end{tabular} &
\begin{tabular}{l}\includegraphics[width=0.198\linewidth]{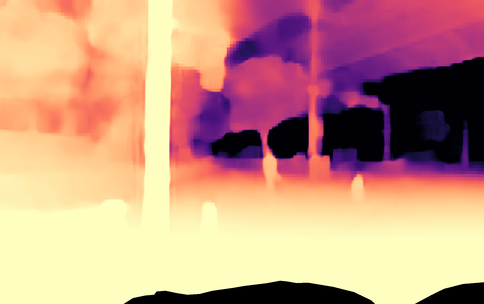}\end{tabular} &
\begin{tabular}{l}\includegraphics[width=0.198\linewidth]{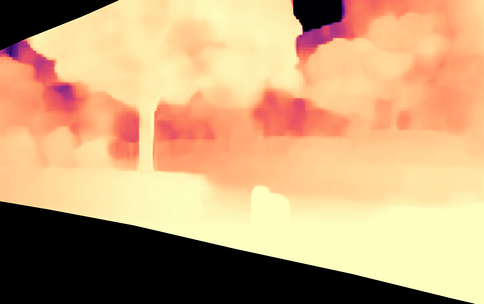}\end{tabular} &
\begin{tabular}{l}\includegraphics[width=0.198\linewidth]{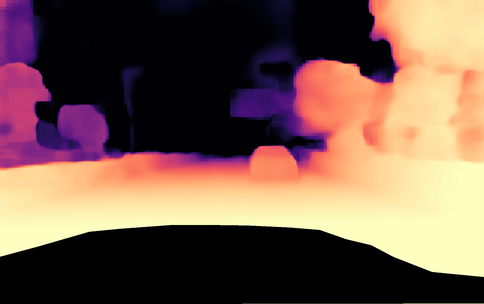}\end{tabular} &
\begin{tabular}{l}\includegraphics[width=0.198\linewidth]{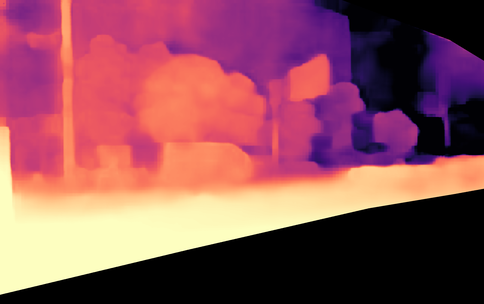}\end{tabular} &
\begin{tabular}{l}\includegraphics[width=0.198\linewidth]{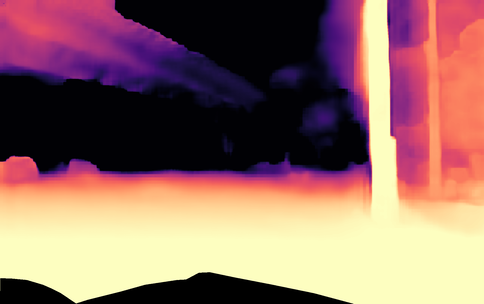}\end{tabular} \\
&
\begin{tabular}{l}\includegraphics[width=0.198\linewidth]{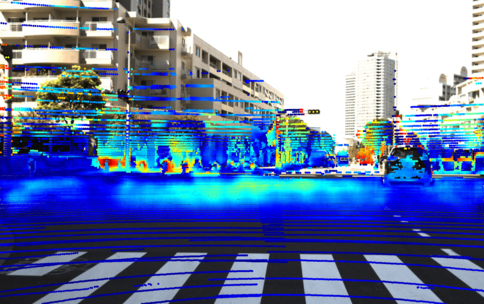}\end{tabular} &
\begin{tabular}{l}\includegraphics[width=0.198\linewidth]{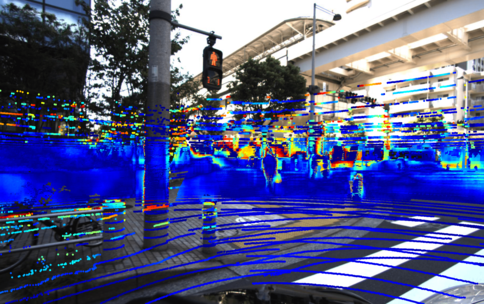}\end{tabular} &
\begin{tabular}{l}\includegraphics[width=0.198\linewidth]{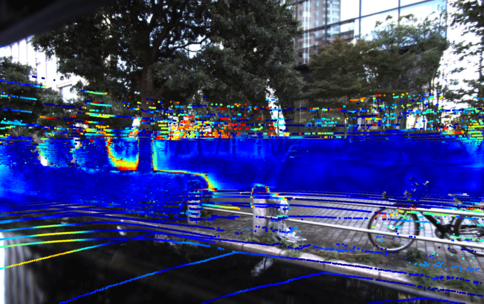}\end{tabular} &
\begin{tabular}{l}\includegraphics[width=0.198\linewidth]{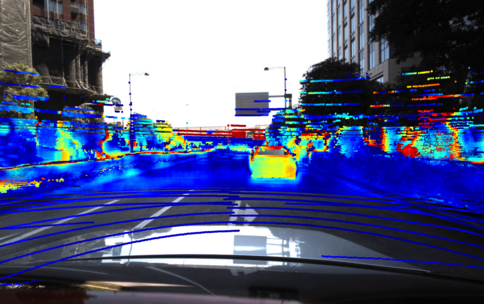}\end{tabular} &
\begin{tabular}{l}\includegraphics[width=0.198\linewidth]{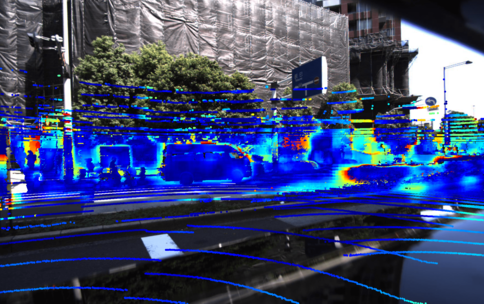}\end{tabular} &
\begin{tabular}{l}\includegraphics[width=0.198\linewidth]{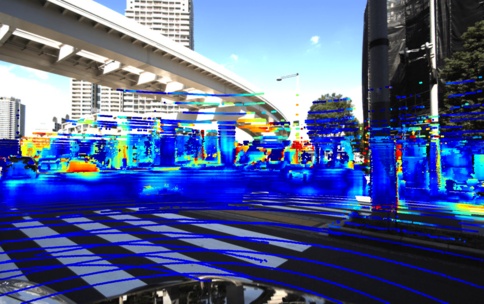}\end{tabular} \\

\end{tabular}}
\vspace{-0.2cm}
\caption{Qualitative comparison of predicted surrounding depth on DDAD dataset~\cite{guizilini2020ddad_packnet}. 
Given the input surrounding images (the top row), we show the visualized depth maps and depth errors of SurroundDepth~\cite{wei2023surrounddepth} and M$^2$Depth. The depth maps are visualized in the range of [$0$, $50m$]. Our method is able to produce more accurate depth with less error and sharper depth edge across multiple cameras.
}
\vspace{-0.2cm}
\label{fig:ddad_error_map}
\end{figure*}

As shown in~\cref{tab:all_results}, M$^2$Depth achieves significant improvement on all metrics compared with existing methods when tested in a similar setting.
To be specific, our method outperforms the SOTA method of single-frame surrounding depth estimation, SurroundDepth~\cite{wei2023surrounddepth}, by 12.02\% on ${Abs. \ Rel.}$ and 13.38\% on ${Sq. \ Rel.}$, indicating that our usage of spatial-temporal volumes substantially improves the depth quality.
We also compare the visualization results of M$^2$Depth and SurroundDepth in~\cref{fig:ddad_error_map}, where the estimated surrounding depth and depth errors show our method produces more accurate and consistent depth predictions among multiple cameras in challenging scenarios.
In \cref{tab:each_camera}, we show the per-camera evaluation results on DDAD. In terms of $Abs. Rel.$, our method is able to predict more accurate depth in nearly all cameras, demonstrating the superior performance of M$^2$Depth.

\paragraph{Results on nuScenes}
In~\cref{tab:all_results}, we evaluate the proposed M$^2$Depth on the evaluation set of nuScenes dataset~\cite{caesar2020nuscenes}, where the quantitative results show our method significantly outperforms the existing method in terms of multiple metrics in similar setting.
Compared with R3D3~\cite{schmied2023r3d3} that use 5 frames as input, our method utilizes only 2 frames and achieves comparable performance on ${Abs. \ Rel.}$ and superior performance on other metrics.
As the test data in nuScenes dataset~\cite{caesar2020nuscenes} is more challenging than DDAD dataset~\cite{guizilini2020ddad_packnet}, the aforementioned results indicate that M$^2$Depth achieves state-of-the-art overall performance.

\paragraph{Memory and Computation Analysis.}
As shown in \cref{tab:each_camera}, compared with R3D3~\cite{schmied2023r3d3} and SurroundDepth~\cite{wei2023surrounddepth}, M$^2$Depth achieves a good balance between overall performance and computational efficiency. 
According to the results, our method consumes much less $memory$ and $FLOPs$ than R3D3~\cite{schmied2023r3d3} while achieving competitive performance.

\begin{table}[htb]
\centering
\caption{
Quantitative results of the ablation study on DDAD dataset~\cite{guizilini2020ddad_packnet}.
M.P. stands for mono prior, S.Vol. and T.Vol. indicate the spatial volume and temporal volume, the STF is the proposed spatial-temporal fusion module.
Jointly constructing spatial-temporal cost volumes significantly improves the depth quality compared with the mono prior depth, and the STF further increases the capabilities of M$^2$Depth on nearly all metrics.
}
\vspace{-0.2cm}
\resizebox{\linewidth}{!}{
\begin{tabular}{cccccccccccc}
\toprule
No. & M.P.   & S.Vol. & T.Vol. & STF & Abs. Rel. $\downarrow$        & Sq. Rel. $\downarrow$       & RMSE $\downarrow$           & RMSE log $\downarrow$      & $\delta < 1.25$ $\uparrow$  & $\delta < 1.25^2$ $\uparrow$ & $\delta < 1.25^3$ $\uparrow$ \\
\midrule
1   & $\checkmark$    &                &                 &               & 0.216   & 3.758  & 13.200 & 0.338    & 0.686 & 0.863 & 0.929  \\
2   & $\checkmark$    & $\checkmark$   &                 &               & 0.212   & 3.662  & 12.959 & {0.326}    & 0.696 & 0.872 & 0.936  \\
3   & $\checkmark$    & $\checkmark$   & $\checkmark$    &               & \underline{0.197}   & \underline{3.379}  & \textbf{12.341} & \underline{0.313}    & \underline{0.738} &\textbf{0.886} & \textbf{0.941}  \\
4   & $\checkmark$    & $\checkmark$   & $\checkmark$    & $\checkmark$  & \textbf{0.194}   & \textbf{3.331}  & \underline{12.347} & \textbf{0.311}    & \textbf{0.741} & \textbf{0.886} & \textbf{0.941}  \\
\bottomrule
\end{tabular}}
\label{tab:ablation_tab1}
\vspace{-0.2cm}
\end{table}

\subsection{Ablation Study}
In this section, we conduct ablation studies on DDAD~\cite{guizilini2020ddad_packnet} to analyze the effectiveness of each module of M$^2$Depth.

\vspace{-0.2cm}
\paragraph{Spatial-temporal Volume}
In~\cref{tab:ablation_tab1}, we evaluate the performance using different cost volumes, where the base model uses none of the spatial volume ($S. \ Vol.$), temporal volume ($T. \ Vol.$) and STF module.
By fusing the spatially adjacent views, $S. \ Vol.$ improves 2.55\% on the $Sq.\  Rel.$ metric and 3.55\% on the $RMSE \ log$ metric.
When further injecting the temporal information into cost volumes, the $T. \ Vol.$ achieves more than 7\% improvement on $Abs. \ Rel$ and $Sq. \ Rel$.
The aforementioned results show that integrating the spatial-temporal information is able to significantly strengthen the depth quality.

\vspace{-0.2cm}
\paragraph{Spatial-temporal Fusion}
Compared with directly using features that warp from previous frames or adjacent views, the proposed STF module fuses the volume features within the spatial-temporal domain, where the updated feature integrates the global information with spatial- and temporal- features and consequently strengthens the feature expressiveness. Compared to only using information from a single domain, our volumes achieve better performance.

\begin{wraptable}{r}{0.55\textwidth}
    \centering
    \vspace{-0.5cm}
    \caption{Ablation studies of $\mathcal{L}_{\mathrm{edge}}$, MFF and D. D. on DDAD dataset~\cite{guizilini2020ddad_packnet}, where $\mathcal{L}_{\mathrm{edge}}$ indicates the depth edge loss, MFF stands for multi-grained feature fusion module, D. D. represents the depth decoding with SAM features.}
    \resizebox{\linewidth}{!}{
    \begin{tabular}{ccccccc}
    \toprule
    $\mathcal{L}_{\mathrm{edge}}$ & MFF  &   D. D.  & Abs. Rel. $\downarrow$   & Sq. Rel. $\downarrow$  & RMSE $\downarrow$  & $\delta < 1.25$ $\uparrow$ \\
    \midrule
                 &              &              & 0.194   & 3.331  & 12.347     & 0.741   \\
     $\checkmark$ &              &              & 0.192   & 3.224  & 12.447     & 0.741   \\
     $\checkmark$ & $\checkmark$ &              & \underline{0.188} & \underline{3.032} & {12.213} & \underline{0.748} \\
     $\checkmark$ &              & $\checkmark$ & 0.191   & 3.262  & \underline{12.175}     & \underline{0.748} \\
     $\checkmark$ & $\checkmark$ & $\checkmark$ & \textbf{0.183}   & \textbf{2.920}  & \textbf{11.963}& \textbf{0.756} \\
    \bottomrule
    \end{tabular}}
    \label{tab:ablation_tab2}
\end{wraptable}

\begin{wrapfigure}{r}{0.55\textwidth}
\vspace{-0.4cm}
\centering
\resizebox{\linewidth}{!}{
\begin{tabular}{lccc}
 & Internal Feature & SAM Feature & Fused Feature \\

\rotatebox[origin=c]{90}{Scene-150} &
\begin{tabular}{l}\includegraphics[width=0.4\linewidth]{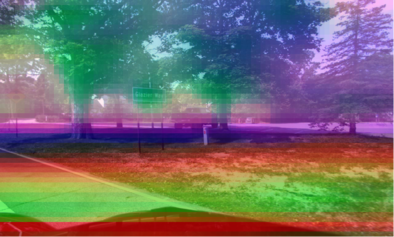}\end{tabular} &
\begin{tabular}{l}\includegraphics[width=0.4\linewidth]{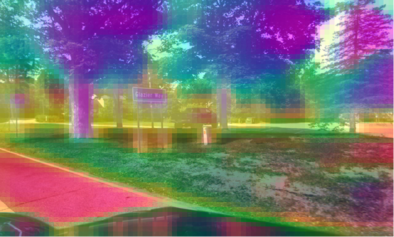}\end{tabular} &
\begin{tabular}{l}\includegraphics[width=0.4\linewidth]{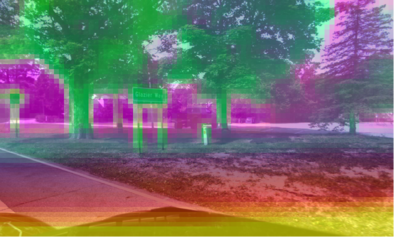}\end{tabular} \\

\rotatebox[origin=c]{90}{Scene-191} &
\begin{tabular}{l}\includegraphics[width=0.4\linewidth]{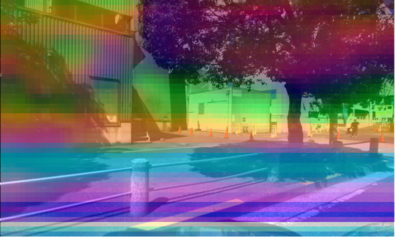}\end{tabular} &
\begin{tabular}{l}\includegraphics[width=0.4\linewidth]{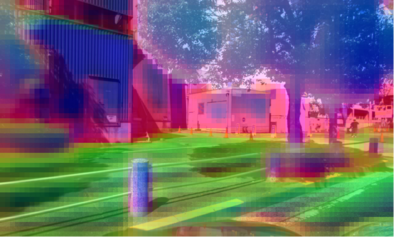}\end{tabular} &
\begin{tabular}{l}\includegraphics[width=0.4\linewidth]{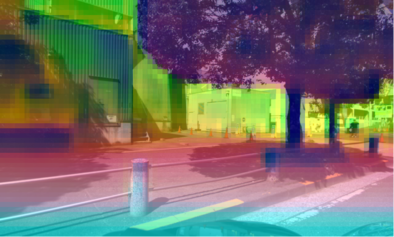}\end{tabular} \\
\end{tabular}}
\caption{Visualization results of different features in M$^2$Depth on DDAD dataset~\cite{guizilini2020ddad_packnet}. The internal feature is from the internal image encoder, the SAM feature is from the frozen SAM encoder~\cite{zhang2023mobile_sam}, and the fused feature is produced by the MFF module.
}
\vspace{-1.0cm}
\label{fig:feature_fusion}
\end{wrapfigure}

\paragraph{Multi-grained Feature Fusion}
As shown in \cref{tab:ablation_tab2}, we conduct the ablation study to evaluate the effectiveness of the MFF module, which enhances feature learning by combining SAM features with internal features. 
According to the quantitative results, introducing MFF into mono prior estimation achieves improvement in nearly all metrics. We also show the visualized features in~\cref{fig:feature_fusion}, where the internal features represent the geometric distance information and the SAM features contain semantic instance information. By combining the internal features and SAM features in latent space, the fused features derive a comprehensive understanding of the surrounding environment.

As mentioned in~\cref{subsec:sam_mono}, we further integrate SAM priors in depth decoding and conduct experimental comparison in~\cref{tab:ablation_tab2}, where the results show that utilizing SAM features in both MFF and depth decoding achieves the best performance of M$^2$Depth.

\paragraph{Edge Loss and Others}
The ablation study of $\mathcal{L}_{\mathrm{edge}}$ is performed in~\cref{tab:ablation_tab2}, where the results indicate $\mathcal{L}_{\mathrm{edge}}$ is able to improve the $Abs. \ Rel.$ and $Sq. \ Rel.$.
We also perform ablation studies for other hyper-parameters and candidate designs, please refer to supplementary materials for more results and analysis.


\section{Conclusion}
\label{sec:conclusion}

\paragraph{Limitation}
Currently, M${^2}$Depth constructs as many volumes as the number of cameras, which consumes a lot of memory when increasing the cameras.
In the future, we'd like to build a unified cost volume to represent the surrounding environment.
\vspace{-0.2cm}
\paragraph{Conclusion}
In this paper, we propose M${^2}$Depth which is designed for the self-supervised two-frame multi-camera metric depth estimation task in autonomous driving.
Different from the previous methods that use single frame or single camera, M${^2}$Depth takes two-frame from multi-camera as inputs and learns to construct spatial-temporal cost volumes, which is the first method to exploit spatial-temporal fusion in constructing cost volumes.
We additionally propose a novel multi-grained feature fusion module to combine the SAM priors with internal features.
Experimental results on two public benchmarks indicate that M${^2}$Depth achieves state-of-the-art performance.

\clearpage
\appendix 
\setcounter{page}{1}

\section{Implementation Details}
\subsection{Depth Decoder}\label{subsec:depth_decoder}
\begin{figure}[!ht]
  \centering
   \includegraphics[width=0.8\linewidth]{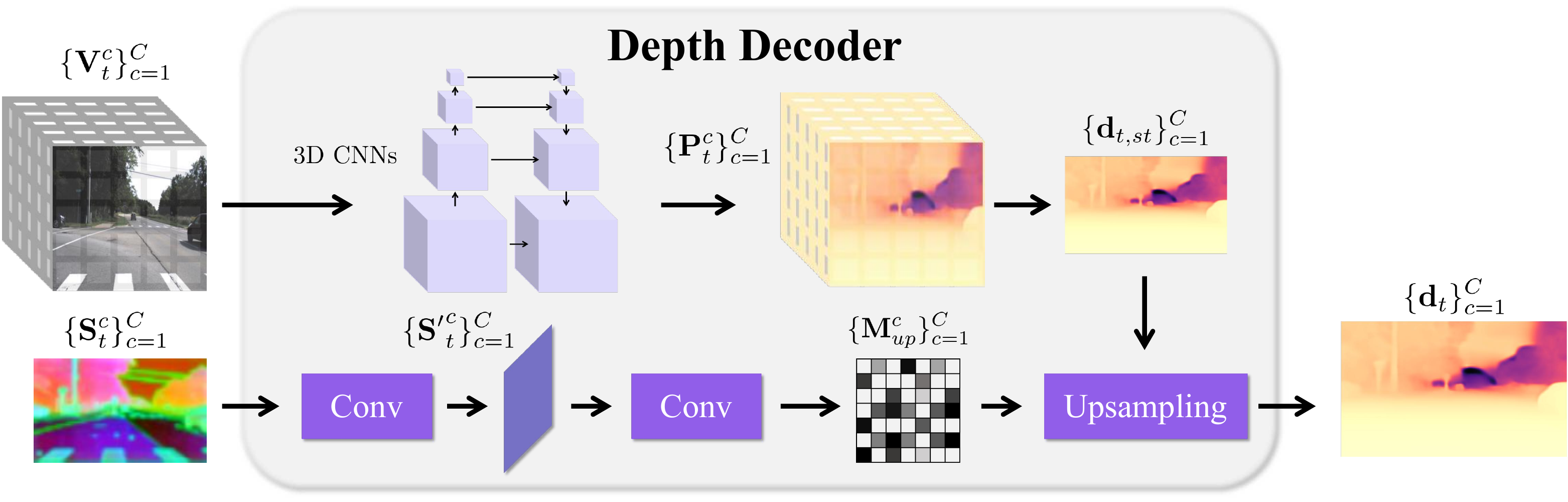}
   \caption{Overview of Depth decoder. Given the spatial-temporal volume $\{\mathbf{V}_t^c\}_{c=1}^C$ and the SAM feature $\{\mathbf{S}_t^c\}_{c=1}^C$ as inputs, we initially compute the spatial-temporal depth $\{\mathbf{d}_{t,st}^c\}_{c=1}^C$. Subsequently, the $\{\mathbf{d}_{t,st}^c\}_{c=1}^C$ is upsampled with the mask $\{\mathbf{M}_{up}^c\}_{c=1}^C$ which are calculated from $\{\mathbf{S}_t^c\}_{c=1}^C$ to procure the final depth $\{\mathbf{d}_{t}^c\}_{c=1}^C$}.
   \label{fig:depth_depth_decoder}
\end{figure}

The detailed structure of the depth decoder is illustrated in ~\cref{fig:depth_depth_decoder}. Given the spatial-temporal volume $\{\mathbf{V}_t^c\}_{c=1}^C$ and the SAM feature $\{\mathbf{S}_t^c\}_{c=1}^C$ from SAM encoder~\cite{zhang2023mobile_sam}, we first transform  $\{\mathbf{V}_t^c\}_{c=1}^C$ into probability volumes$\{\mathbf{P}_t^c\}_{c=1}^C$ by 3D CNNs. Then, we calculate the spatial-temporal depth $\{\mathbf{d}_{t,st}^c\}_{c=1}^C$ using depth samples. Subsequently, we utilize $\{\mathbf{S}_t^c\}_{c=1}^C$ as context features to compute the upsampling mask $\{\mathbf{M}_{up}^c\}_{c=1}^C$. Finally, by integrating $\{\mathbf{M}_{up}^c\}_{c=1}^C$ and $\{\mathbf{d}_{t,st}^c\}_{c=1}^C$, we can obtain the final depth $\{\mathbf{d}_{t}^c\}_{c=1}^C$.
\vspace{-0.5cm}

\vspace{-1.0cm}
\subsection{Adaptive Depth Sample}
\label{subsec:adaptive_sample}
Following the plane sweep paradigm, the selection of depth samples directly affects the depth quality. Previous methods~\cite{ding2022transmvsnet, yao2018mvsnet, gu2020casmvsnet} usually adopt a wide-range sampling strategy for the entire scene, which improves the accuracy of depth estimation to some extent, but also brings a huge computational burden. 

To solve this problem, we propose utilizing the mono depth estimation result as prior information and conducting adaptive sampling in the vicinity of the prior depth. This method not only significantly reduces the computational complexity, but also improves the efficiency of depth estimation.

\begin{wrapfigure}{r}{0.5\textwidth}
  \centering
  \vspace{-0.7cm}
   \includegraphics[width=\linewidth]{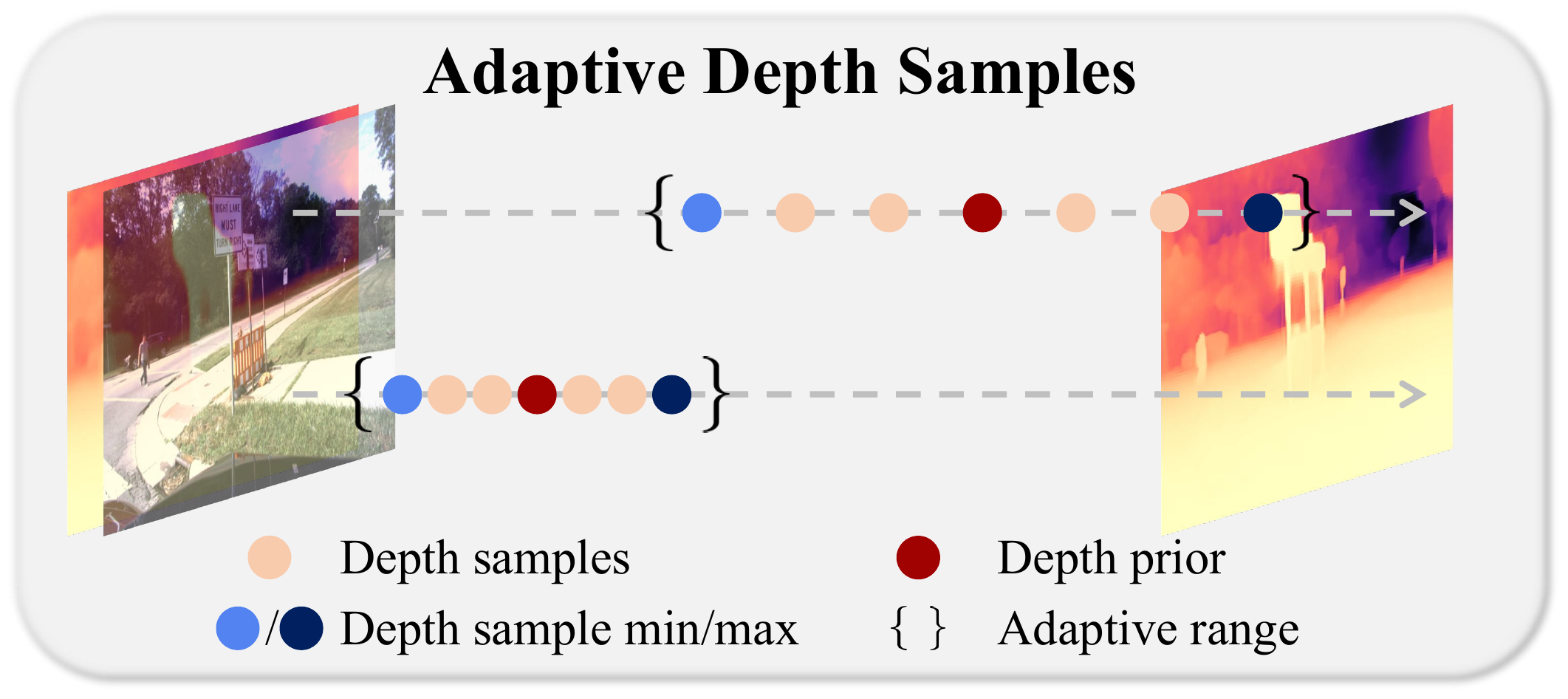}
   \caption{We illustrate the examples of the adaptive depth sample, where the depth range increases for pixels at a farther distance, and conversely, decreases for pixels at a closer proximity.}
   \label{fig:adaptive_range}
  \vspace{-1.0cm}
\end{wrapfigure}
The method of adaptive depth sampling is shown in~\cref{fig:adaptive_range}. Specifically, we determine the range of depth sampling $[\mathbf{d}_{\mathrm{min}}(\mathbf{p}) , \mathbf{d}_{\mathrm{max}}(\mathbf{p})]$ for each pixel $\mathbf{p}$ based on the given depth $\mathbf{d}_{\mathrm{init}}$ and scaling factor $\alpha$ as follow:
\begin{equation}
    \mathbf{d}_{\mathrm{min}}(\mathbf{p}) = \mathbf{d}_{\mathrm{init}}(\mathbf{p}) \div (1 + \alpha), 
    \label{eq:d_range_min}
\end{equation}
\begin{equation}
    \mathbf{d}_{\mathrm{max}}(\mathbf{p}) = \mathbf{d}_{\mathrm{init}}(\mathbf{p}) \times (1 + \alpha), 
    \label{eq:d_range_max}
\end{equation}

It is evident from this formula that the depth range varies with the depth. When the $\mathbf{d}_{\mathrm{init}}(\mathbf{p})$ is large, that is, the object is farther away, the range of depth sampling will increase accordingly; conversely, when the $\mathbf{d}_{\mathrm{init}}(\mathbf{p})$ is small, the range of depth sampling will decrease. This adaptive depth sampling strategy is more in line with the depth distribution of actual scenes, thus effectively improving the quality of depth.

\subsection{Structure-from-Motion Loss}
Through self-supervised photometric loss $\mathcal{L}_{\mathrm{photo}}$, we can effectively supervise the estimated depth and pose. However, during the initial phase of training, obtaining valid projection results is challenging due to insufficient overlap between adjacent cameras, which ultimately renders supervision ineffective. To address this issue, we follow previous methods~\cite{guizilini2022full_surround_monodepth, wei2023surrounddepth} and obtain scale-aware depth through triangulation of adjacent cameras utilizing their camera extrinsics, which serves as pseudo labels for effective supervision. By doing so, we successfully enhance the accuracy of depth and pose estimation by leveraging information from neighboring cameras and extrinsics.

The calculation for $\mathcal{L}_{\mathrm{sfm}}$ is as follows:
\begin{equation}
    \mathcal{L}_{\mathrm{sfm}} = \frac{1}{\left| \mathbb{M} \right|} \sum_{\mathbf{p} \in \mathbb{M}}
    \left| \mathbf{d}(\mathbf{p}) - \mathbf{d}_{\mathrm{sfm}}(\mathbf{p}) \right|_{1},
    \label{eq:l_sfm}
\end{equation}
where $\mathbb{M}$ represents the set of valid pixel $\mathbf{p}$ in pseudo depth labels $\mathbf{d}_{\mathrm{sfm}}$.

\subsection{Evaluation Metrics}
Following in previous work~\cite{wei2023surrounddepth, guizilini2022full_surround_monodepth}, the description of the evaluation metrics we used is as follows:
\begin{equation}
\text{Abs.Rel.: } \frac{1}{|\mathbb{N}|} \sum_{\mathbf{p} \in \mathbb{N}} \frac{\left|\mathbf{d}(\mathbf{p})-\mathbf{d}^{*}(\mathbf{p})\right|}{\mathbf{d}^{*}(\mathbf{p})},
\label{eq:abs_rel}
\end{equation}

\begin{equation}
\text{Sq. Rel.: } \frac{1}{|\mathbb{N}|} \sum_{\mathbf{p} \in \mathbb{N}} \frac{\left\|\mathbf{d}(\mathbf{p})-\mathbf{d}^{*}(\mathbf{p})\right\|^2}{\mathbf{d}^{*}(\mathbf{p})},
\label{eq:sq_rel}
\end{equation}

\begin{equation}
\text{RMSE: } \frac{1}{|\mathbb{N}|} \sqrt{\sum_{\mathbf{p} \in \mathbb{N}}\left\|\mathbf{d}(\mathbf{p})-\mathbf{d}^*(\mathbf{p})\right\|^2},
\label{eq:rmse}
\end{equation}

\begin{equation}
\text{RMSE log: } \frac{1}{|\mathbb{N}|} \sqrt{\sum_{\mathbf{p} \in \mathbb{N}}\left\|\mathrm{log}\mathbf{d}(\mathbf{p})-\mathrm{log}\mathbf{d}^{*}(\mathbf{p})\right\|^2},
\label{eq:rmse_log}
\end{equation}

\begin{equation}
\delta<n\text{: fraction of } d  \in \mathbf{d} \text{ for which } \max \left(\frac{d}{d^*}, \frac{d^*}{d}\right)<n,
\label{eq:delta}
\end{equation}
where $\mathbf{d}$ and $\mathbf{d^*}$ indicate the predicted depth and ground-truth depth respectively. $\mathbb{N}$ indicates the all valid pixels $\mathbf{p}$ in $\mathbf{d^*}$.

\section{Computation Analysis}
In Tab. ~\cref{tab:module_computation}, we show the computation cost of each module. It can be observed that the cost volume construction and fusion occupy
a high proportion of memory and time, as the grid sample operation is well known to be time-consuming. Reducing the runtime in V.C.F is an important future work.

\begin{table}[htp]
\centering
\caption{
Computation analysis of each module: Pose Branch (Pose), Image Encoder (I.E.), SAM Encoder (S.E.), Prior Decoder (P.D.), Volume Construct \& Fusion (V.C.F.), Depth Decoder (D.D.).
Experiments are performed on V100.
}
\resizebox{0.7\linewidth}{!}{
\begin{tabular}{lcccccccc}
\toprule
            & Pose  & I.E. & S.E. & MFF    & P.D. & V.C.F. & D.D. \\
\midrule
Memory(MB)  & 139.20      & 139.07        & 173.03      & 51.10  & 105.39        & 397.12                      & 196.33         \\
Percent(\%) & 11.59\%     & 11.58\%       & 14.40\%     & 4.25\% & 8.77\%        & 33.06\%                     & 16.34\%        \\
Time(ms)    & 39.33       & 3.35          & 20.65       & 3.58   & 1.39          & 216.35                      & 2.34           \\
Percent(\%) & 13.71\%     & 1.17\%        & 7.20\%      & 1.25\% & 0.48\%        & 75.39\%                     & 0.81\%         \\
\bottomrule
\end{tabular}}
\label{tab:module_computation}
\end{table}

\section{Ablation Study}

\paragraph{Design of Pose Estimation}
\cref{tab:pose_branch} shows that the $\textit{Front Camera}$ ($\textit{F. Cam.}$) can achieve better results. We take the previous method~\cite{wei2023surrounddepth} which concatenates surrounding views to directly predict the ego pose as the baseline $\textit{Concat Camera}$ ($\textit{C. Cam.}$). Experiments indicate that the method $\textit{F. Cam.}$, which predict the pose of front-view camera $\mathbf{P}^0_{t \to t-1}$ and then derive the ego pose $\mathbf{P}_{t \to t-1}$, is more effective.

\begin{table}[!t]
        \begin{minipage}[t]{0.49\textwidth}
            \centering
            \caption{Ablation study on the design of pose estimation module comparison. Experiments demonstrate that the method, which utilizes the front-view camera to estimate the front-view pose and subsequently infer the ego pose, is well-suited for our depth estimation network and embodies its effectiveness. (\textbf{Bold} figures indicate the best and \underline{underlined} figures indicate the second best) }
            \resizebox{\linewidth}{!}{
            \begin{tabular}{lcccccc}
            \toprule
            \textbf{Method}  & Abs. Rel. & Sq. Rel.  & RMSE   & RMSE log  & $\delta < 1.25$  \\
            \midrule
            C. Cam.  & \underline{0.189}   & \underline{2.942}  & \underline{12.239} & \underline{0.309}    & \underline{0.732}   \\
            F. Cam. & \textbf{0.183}   & \textbf{2.920} & \textbf{11.963} & \textbf{0.299}  & \textbf{0.756}  \\
            \bottomrule
            \end{tabular}}
            \label{tab:pose_branch}
        \end{minipage}
        \hspace{0.02\textwidth}
        \begin{minipage}[t]{0.49\textwidth}
            \centering
            \caption{Designs of feature fusion module comparison. We train MFF as described in the main paper and train the VFF module which fuses the internal feature and SAM feature through direct addition. Experimental results demonstrate that our design effectively integrates diverse-grained features, thereby significantly enhancing the quality of depth estimation. (\textbf{Bold} figures indicate the best and \underline{underlined} figures indicate the second best) }
            \resizebox{\linewidth}{!}{
            \begin{tabular}{lcccccc}
            \toprule
            \textbf{Method}  & Abs. Rel. & Sq. Rel.  & RMSE   & RMSE log  & $\delta < 1.25$ \\
            \midrule
            Base & {0.191}   & {3.262}  & \underline{12.175} &\underline{0.305}    & \underline{0.748}   \\
            VFF  & \underline{0.185} & \underline{3.044} & {12.209} &{0.307} & {0.746} \\
            MFF  & \textbf{0.183} & \textbf{2.920}  & \textbf{11.963} & \textbf{0.299}  & \textbf{0.756} \\
            \bottomrule
            \end{tabular}}
            \label{tab:ablation_mff}
        \end{minipage}
\end{table}

\paragraph{Design of Multi-grained Feature Fusion Module}
In \cref{tab:ablation_mff}, we evaluate the performance of different feature fusion methods in mono prior estimation. Specifically, we compare the base model, which does not utilize the MFF module, against the multi-grained feature fusion (MFF) module and the vanilla feature fusion (VFF) module that blends SAM features with internal features through simple addition.
The results presented in \cref{tab:ablation_mff} demonstrate that the incorporation of SAM features notably elevates the quality of depth estimation outcomes.
Comparing the MFF module with the VFF module, our multi-grained feature fusion module exhibits superior performance in fusing internal features with fine-grained semantic information, thereby further augmenting the precision of depth estimation.

\begin{table}[tb]
        \begin{minipage}[t]{0.49\textwidth}
            \centering
            \caption{Designs of depth decoder comparison. We train SAM Refine ($\textit{S. Refine}$) as described in the main paper and train Vanilla Refine ($\textit{V. Refine}$) using the context feature from FPN~\cite{lin2017fpn}. We evaluate both the network on DDAD and the experiments show that SAM Refine effectively enhances depth quality. (\textbf{Bold} figures indicate the best and \underline{underlined} figures indicate the second best) }
            \resizebox{\linewidth}{!}{
            \begin{tabular}{lcccccc}
            \toprule
            \textbf{Method}  & Abs. Rel. & Sq. Rel.  & RMSE   & RMSE log  & $\delta < 1.25$ \\
            \midrule
            Base  & \underline{0.192}   & \textbf{3.224}  & 12.447 & \underline{0.312}    & \underline{0.741}   \\
            V. Refine & 0.196   & 3.313  & \underline{12.366} & 0.313  & 0.734  \\
            S. Refine & \textbf{0.191} & \underline{3.262}  & \textbf{12.175} & \textbf{0.305}  & \textbf{0.748} \\
            \bottomrule
            \end{tabular}}
            \label{tab:ablation_refine}
        \end{minipage}
        \hspace{0.02\textwidth}
        \begin{minipage}[t]{0.49\textwidth}
            \centering
            \caption{Designs of depth sample comparison. We train Adaptive Sample ($\textit{A. Sample}$), Vanilla Sample ($\textit{V. Sample}$) and Fixed Sample ($\textit{F. Sample}$) with 16 samples. We evaluate both the network on DDAD and the experiments show that using adaptive methods yields better results. (\textbf{Bold} figures indicate the best and \underline{underlined} figures indicate the second best) }
            \resizebox{\linewidth}{!}{
            \begin{tabular}{lcccccc}
            \toprule
            \textbf{Method}  & Abs. Rel. & Sq. Rel.  & RMSE   & RMSE log  & $\delta < 1.25$ \\
            \midrule
            V. Sample & 0.362  & 5.932  & 14.891 & 0.422  & 0.534  \\
            F. Sample & \underline{0.195}  & \underline{3.054}  & \underline{12.362} & \underline{0.309}  & \underline{0.721}  \\
            A. Sample & \textbf{0.183}   & \textbf{2.920} & \textbf{11.963} & \textbf{0.299}  & \textbf{0.756}  \\
            \bottomrule
            \end{tabular}}
            \label{tab:ablation_adaptive}
        \end{minipage}
\end{table}
\paragraph{Design of Depth Decoder}
For \cref{tab:ablation_refine}, we train two variants of our depth decoder: Vanilla Refine ($\textit{V. Refine}$) and SAM Refine ($\textit{S. Refine}$). The former utilizes context features from FPN~\cite{lin2017fpn}, whereas the latter employs context features from the SAM encoder~\cite{kirillov2023segment_anything}. Through evaluation on the DDAD dataset, $\textit{S. Refine}$ attains superior results. The results show that the network necessitates the integration of more fine-grained information to enhance depth refinement. When compared to FPN features, which encompass feature-matching information, SAM features are deemed more suitable.

\paragraph{Adaptive Depth Sample}
In ~\cref{tab:ablation_adaptive}, we perform a comparison between the adaptive depth samples as described in the main paper ($\textit{A. Sample}$), the fixed depth samples within a fixed depth sampling range ($\textit{F. Sample}$), the vanilla depth sample within the entire space ($\textit{V. Sample}$). The experimental results consistently show that the adaptive method yields better outcomes.

\paragraph{Number of Bins}
We conduct an ablation study against the number of bins on DDAD~\cite{guizilini2020ddad_packnet} dataset, and the results are shown in \cref{tab:bins}. Our results demonstrate that increasing the quantity of bins does not significantly enhance the quality of depth. This indicates that the utilization of adaptive depth samples effectively contributes to improving computational efficiency.

\begin{table}[tb]
        \begin{minipage}[t]{0.49\textwidth}
            \centering
            \caption{Ablation study on number of bins. We compare the influence of the different number of bins used to train the network. (\textbf{Bold} figures indicate the best and \underline{underlined} figures indicate the second best) }
            \resizebox{\linewidth}{!}{
            \begin{tabular}{lcccccc}
            \toprule
            \textbf{Bins} & Abs. Rel. & Sq. Rel. & RMSE   & $\delta < 1.25$  & Memory(MB)\\
            \midrule
            8    & \underline{0.195}     & \underline{3.316}    & \underline{12.349} & \underline{0.740} & \textbf{3483} \\
            16   & \textbf{0.194}     & 3.331    & \textbf{12.347} & \textbf{0.741} & \underline{3853} \\
            32   & 0.200     & \textbf{3.264}    & 12.491 & 0.724 & 4751 \\  
            \bottomrule
            \end{tabular}}
            \label{tab:bins}
        \end{minipage}
        \hspace{0.02\textwidth}
        \begin{minipage}[t]{0.49\textwidth}
            \centering
            \caption{Ablation study on number of frames. The experimental results demonstrate that our method achieves highly competitive results with just two frames. (\textbf{Bold} figures indicate the best and \underline{underlined} figures indicate the second best) }
            \resizebox{\linewidth}{!}{
            \begin{tabular}{cccccccc}
            \toprule
            \textbf{Frames}  & Abs. Rel. & Sq. Rel.  & RMSE   & RMSE log  & $\delta < 1.25$ \\
            \midrule
            (-1, 0) & \textbf{0.183}   & \underline{2.920} & \textbf{11.963} & \textbf{0.299}  & \textbf{0.756}  \\
            (-2, -1, 0) & \underline{0.185}   & {2.956} & \underline{12.100} & \underline{0.301}  & \underline{0.747}  \\
            (-3, -2, -1, 0) & {0.186}   & \textbf{2.911} & {12.185} & {0.303}  & {0.740}  \\
            \bottomrule
            \end{tabular}}
            \label{tab:more_frames}
        \end{minipage}
\end{table}
\vspace{-0.2cm}

\paragraph{More Frames}
We conduct a multi frames experiment using multiple frames (2 frames, 3 frames, 4 frames) as inputs for depth estimation. ~\cref{tab:more_frames} reveals that increasing the number of frames does not necessarily improve depth accuracy. As our method is not specifically designed to handle sequence data, increasing the input frames does not effectively contribute new information. Notably, employing just two frames is sufficient to produce commendable results.

\section{Visualized}
\vspace{-0.2cm}
\subsection{SAM Feature Enhanced Depth}
\begin{figure}[htb]
\setlength\tabcolsep{0.5 pt}
\centering
\resizebox{0.8\linewidth}{!}{
\begin{tabular}{lcccc}
& Images & Prior w/o SAM & Prior with SAM & Full Model \\

\rotatebox[origin=c]{90}{Scene-150} &
\begin{tabular}{l}\includegraphics[width=0.25\linewidth]{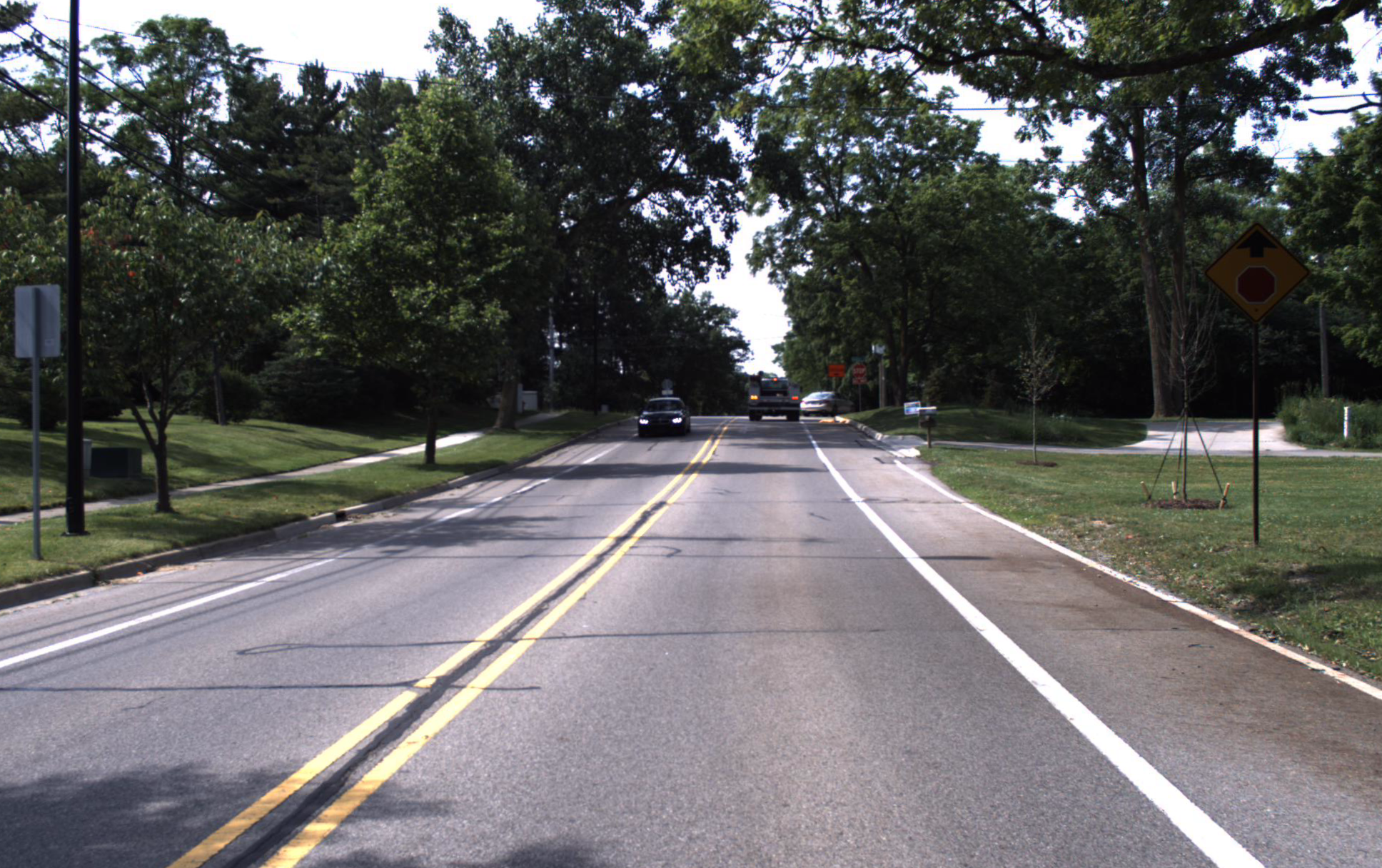}\end{tabular} &
\begin{tabular}{l}\includegraphics[width=0.25\linewidth]{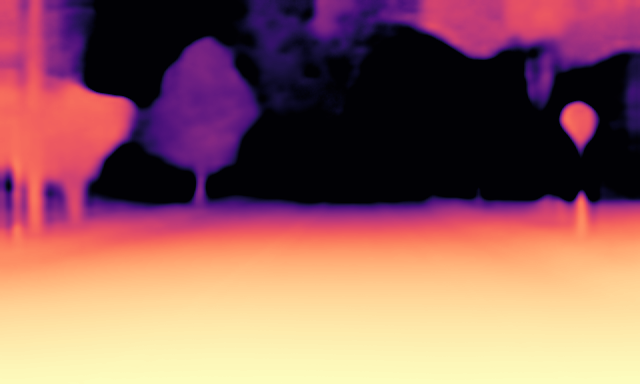}\end{tabular} &
\begin{tabular}{l}\includegraphics[width=0.25\linewidth]{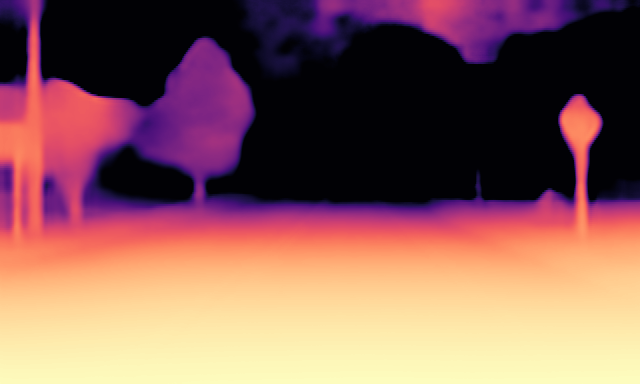}\end{tabular} &
\begin{tabular}{l}\includegraphics[width=0.25\linewidth]{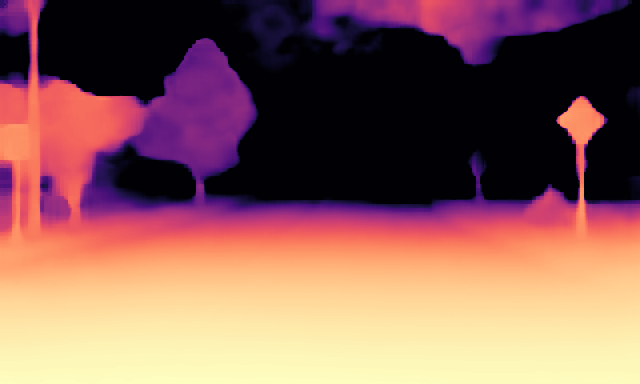}\end{tabular} \\

\rotatebox[origin=c]{90}{Scene-187} &
\begin{tabular}{l}\includegraphics[width=0.25\linewidth]{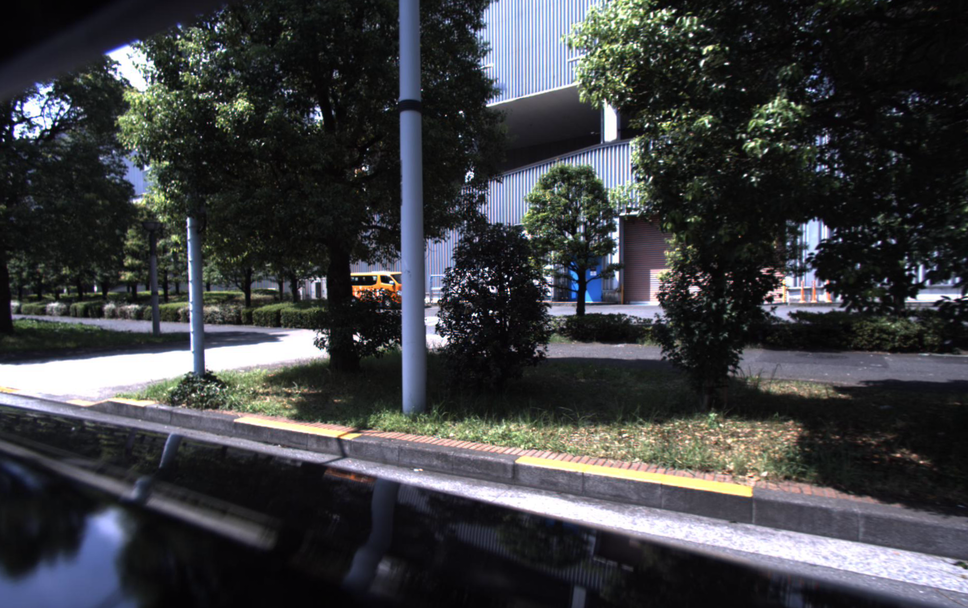}\end{tabular} &
\begin{tabular}{l}\includegraphics[width=0.25\linewidth]{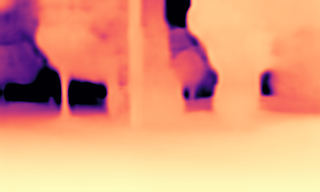}\end{tabular} &
\begin{tabular}{l}\includegraphics[width=0.25\linewidth]{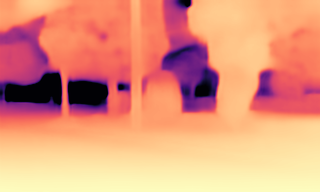}\end{tabular} &
\begin{tabular}{l}\includegraphics[width=0.25\linewidth]{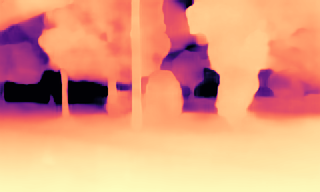}\end{tabular} \\
\end{tabular}}
\vspace{-0.2cm}
\caption{Visualization of produced depth results on DDAD dataset~\cite{guizilini2020ddad_packnet}. It can be observed clearly that consistency within instances and discrimination between different instances for both depths has improved.}
\vspace{-0.2cm}
\label{fig:feature_fusion_depth}
\end{figure}

As shown in ~\cref{fig:feature_fusion_depth}, integrating SAM features gets a notable enhancement in both the depth prior and the final depth, particularly evident at the edges of the instance.

\vspace{-0.2cm}
\subsection{More Depth Results}
\begin{figure*}[t]
\setlength\tabcolsep{0.5 pt}
\centering
\scalebox{0.8}{
\begin{tabular}{lcccccc}

& Front & F.Left & B.Left & Back & B.Right & F.Right \\
\rotatebox[origin=c]{90}{Input} &
\begin{tabular}{l}\includegraphics[width=0.198\linewidth]{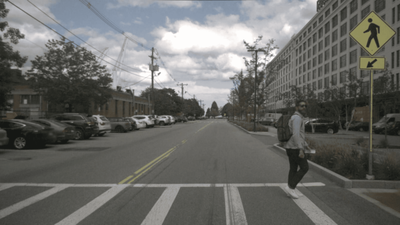}\end{tabular} &
\begin{tabular}{l}\includegraphics[width=0.198\linewidth]{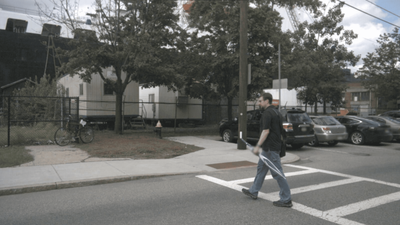}\end{tabular} &
\begin{tabular}{l}\includegraphics[width=0.198\linewidth]{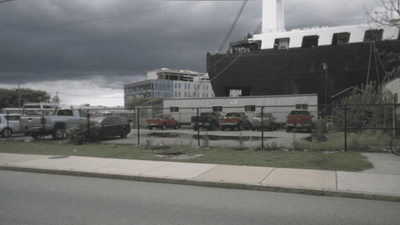}\end{tabular} &
\begin{tabular}{l}\includegraphics[width=0.198\linewidth]{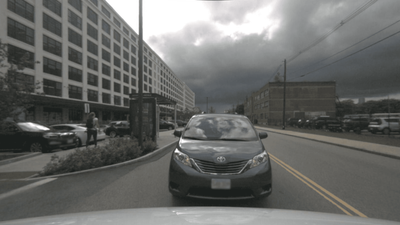}\end{tabular} &
\begin{tabular}{l}\includegraphics[width=0.198\linewidth]{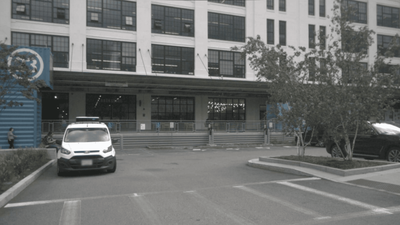}\end{tabular} &
\begin{tabular}{l}\includegraphics[width=0.198\linewidth]{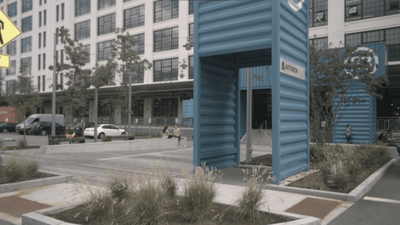}\end{tabular} \\

\rotatebox[origin=c]{90}{SD~\cite{wei2023surrounddepth}} &
\begin{tabular}{l}\includegraphics[width=0.198\linewidth]{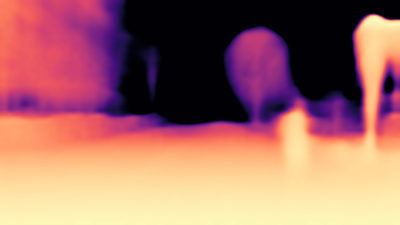}\end{tabular} &
\begin{tabular}{l}\includegraphics[width=0.198\linewidth]{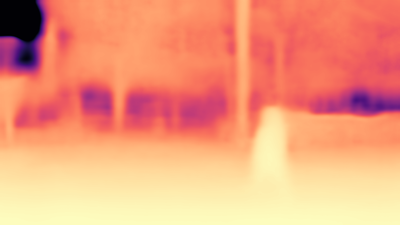}\end{tabular} &
\begin{tabular}{l}\includegraphics[width=0.198\linewidth]{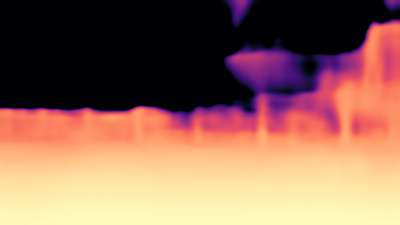}\end{tabular} &
\begin{tabular}{l}\includegraphics[width=0.198\linewidth]{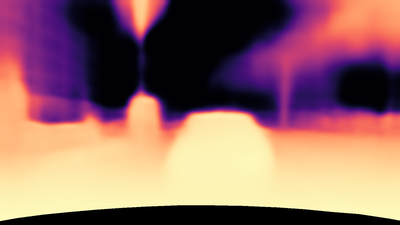}\end{tabular} &
\begin{tabular}{l}\includegraphics[width=0.198\linewidth]{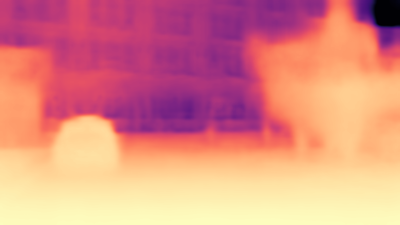}\end{tabular} &
\begin{tabular}{l}\includegraphics[width=0.198\linewidth]{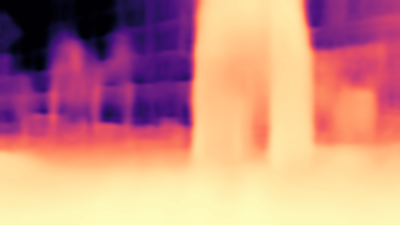}\end{tabular} \\

\rotatebox[origin=c]{90}{Ours} &
\begin{tabular}{l}\includegraphics[width=0.198\linewidth]{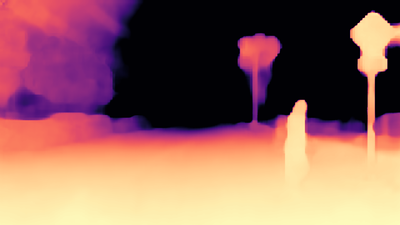}\end{tabular} &
\begin{tabular}{l}\includegraphics[width=0.198\linewidth]{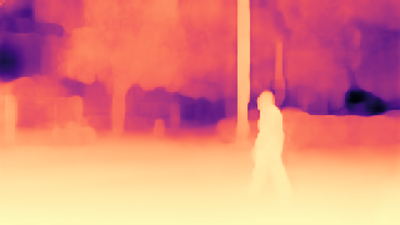}\end{tabular} &
\begin{tabular}{l}\includegraphics[width=0.198\linewidth]{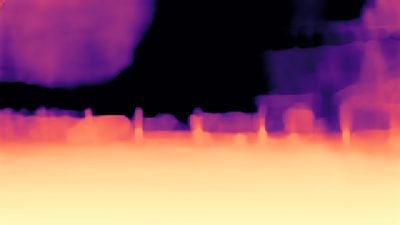}\end{tabular} &
\begin{tabular}{l}\includegraphics[width=0.198\linewidth]{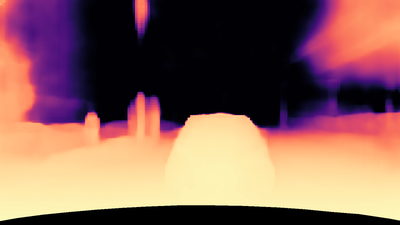}\end{tabular} &
\begin{tabular}{l}\includegraphics[width=0.198\linewidth]{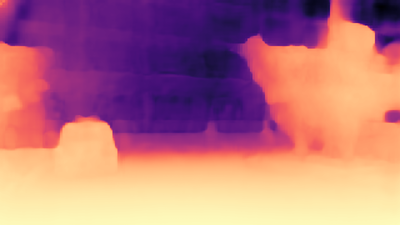}\end{tabular} &
\begin{tabular}{l}\includegraphics[width=0.198\linewidth]{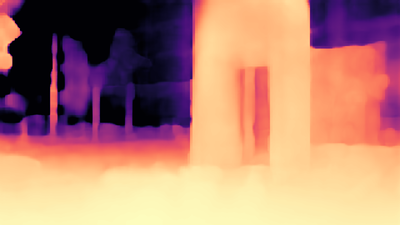}\end{tabular} \\

\midrule

\rotatebox[origin=c]{90}{Input} &
\begin{tabular}{l}\includegraphics[width=0.198\linewidth]{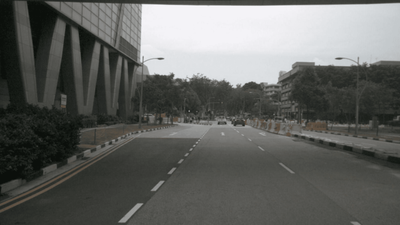}\end{tabular} &
\begin{tabular}{l}\includegraphics[width=0.198\linewidth]{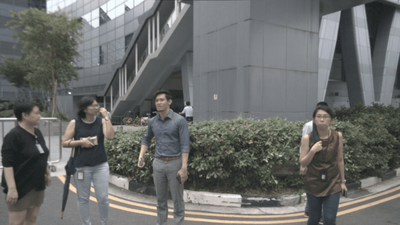}\end{tabular} &
\begin{tabular}{l}\includegraphics[width=0.198\linewidth]{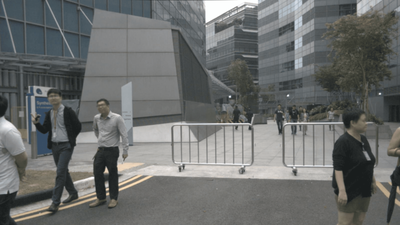}\end{tabular} &
\begin{tabular}{l}\includegraphics[width=0.198\linewidth]{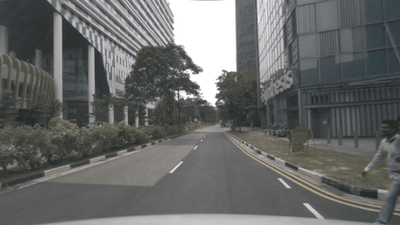}\end{tabular} &
\begin{tabular}{l}\includegraphics[width=0.198\linewidth]{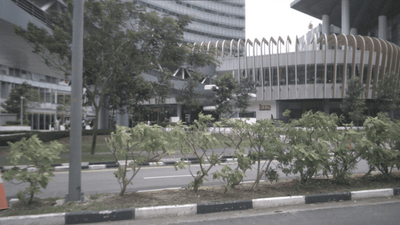}\end{tabular} &
\begin{tabular}{l}\includegraphics[width=0.198\linewidth]{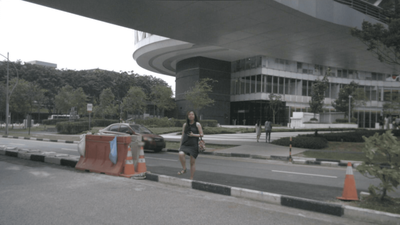}\end{tabular} \\

\rotatebox[origin=c]{90}{SD~\cite{wei2023surrounddepth}} &
\begin{tabular}{l}\includegraphics[width=0.198\linewidth]{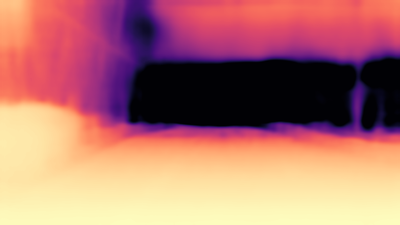}\end{tabular} &
\begin{tabular}{l}\includegraphics[width=0.198\linewidth]{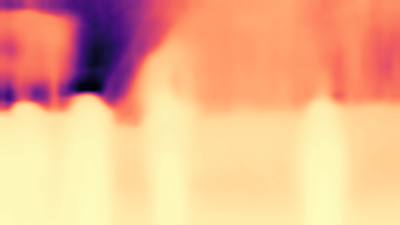}\end{tabular} &
\begin{tabular}{l}\includegraphics[width=0.198\linewidth]{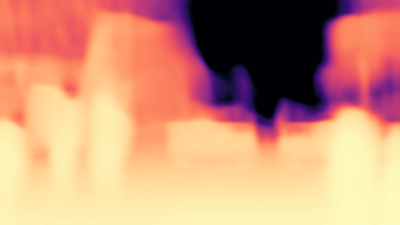}\end{tabular} &
\begin{tabular}{l}\includegraphics[width=0.198\linewidth]{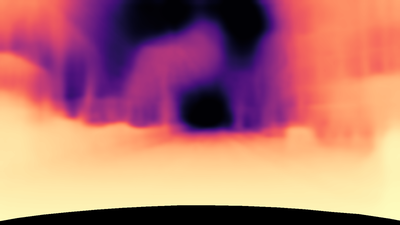}\end{tabular} &
\begin{tabular}{l}\includegraphics[width=0.198\linewidth]{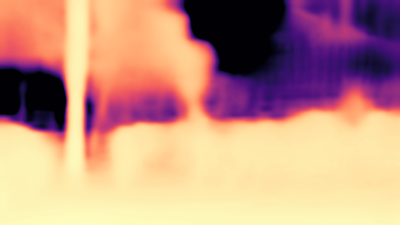}\end{tabular} &
\begin{tabular}{l}\includegraphics[width=0.198\linewidth]{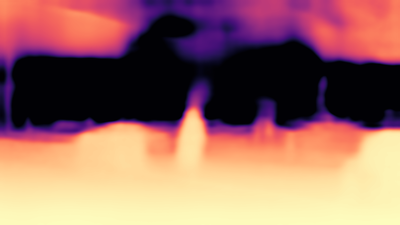}\end{tabular} \\

\rotatebox[origin=c]{90}{Ours} &
\begin{tabular}{l}\includegraphics[width=0.198\linewidth]{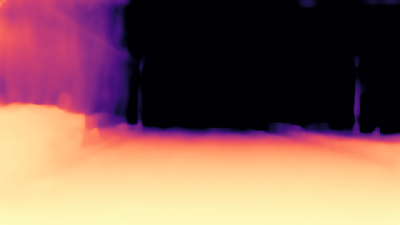}\end{tabular} &
\begin{tabular}{l}\includegraphics[width=0.198\linewidth]{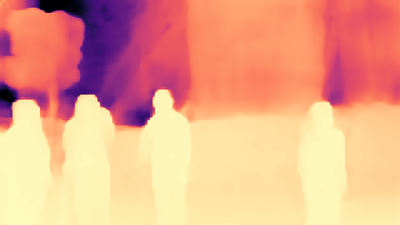}\end{tabular} &
\begin{tabular}{l}\includegraphics[width=0.198\linewidth]{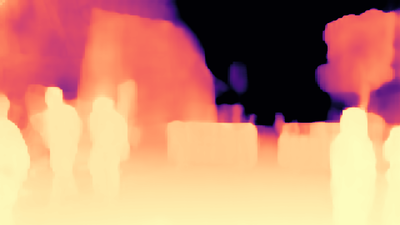}\end{tabular} &
\begin{tabular}{l}\includegraphics[width=0.198\linewidth]{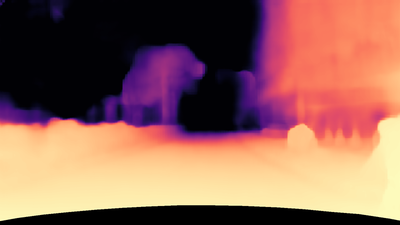}\end{tabular} &
\begin{tabular}{l}\includegraphics[width=0.198\linewidth]{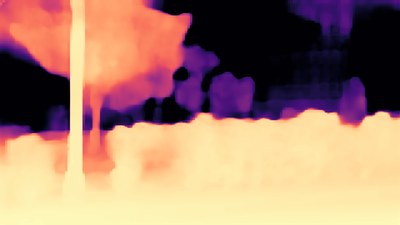}\end{tabular} &
\begin{tabular}{l}\includegraphics[width=0.198\linewidth]{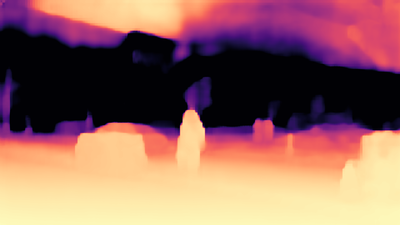}\end{tabular} \\

\midrule

\rotatebox[origin=c]{90}{Input} &
\begin{tabular}{l}\includegraphics[width=0.198\linewidth]{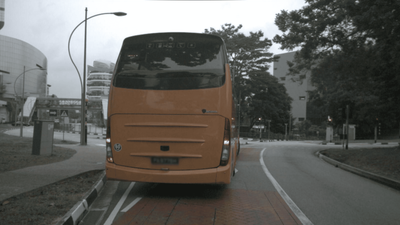}\end{tabular} &
\begin{tabular}{l}\includegraphics[width=0.198\linewidth]{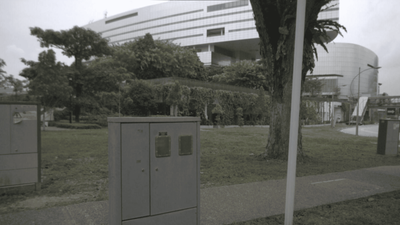}\end{tabular} &
\begin{tabular}{l}\includegraphics[width=0.198\linewidth]{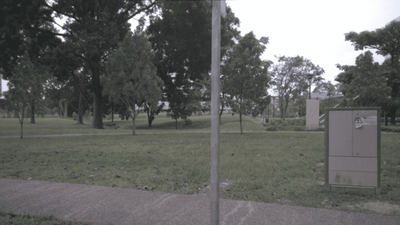}\end{tabular} &
\begin{tabular}{l}\includegraphics[width=0.198\linewidth]{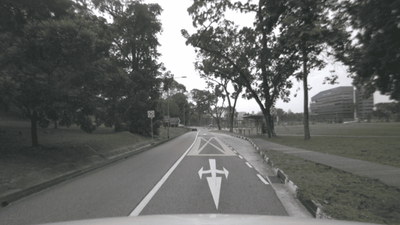}\end{tabular} &
\begin{tabular}{l}\includegraphics[width=0.198\linewidth]{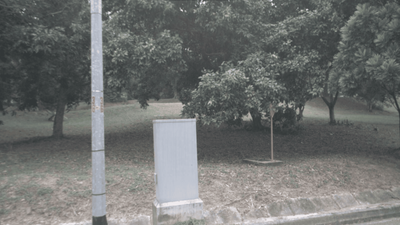}\end{tabular} &
\begin{tabular}{l}\includegraphics[width=0.198\linewidth]{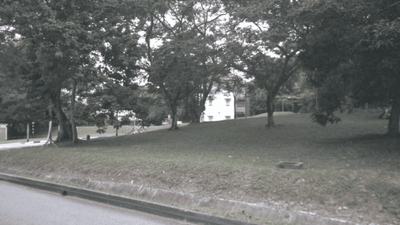}\end{tabular} \\

\rotatebox[origin=c]{90}{SD~\cite{wei2023surrounddepth}} &
\begin{tabular}{l}\includegraphics[width=0.198\linewidth]{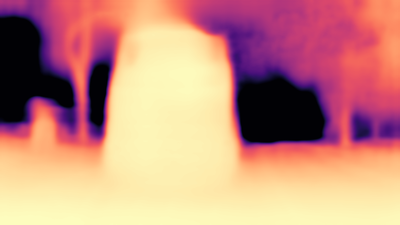}\end{tabular} &
\begin{tabular}{l}\includegraphics[width=0.198\linewidth]{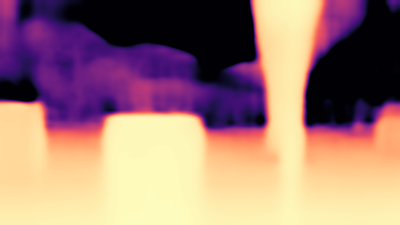}\end{tabular} &
\begin{tabular}{l}\includegraphics[width=0.198\linewidth]{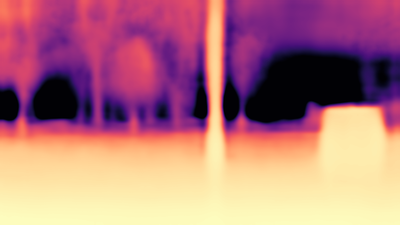}\end{tabular} &
\begin{tabular}{l}\includegraphics[width=0.198\linewidth]{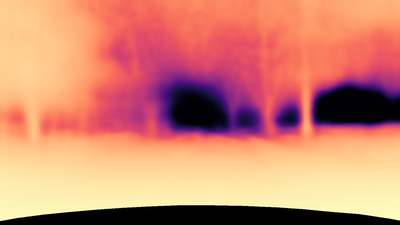}\end{tabular} &
\begin{tabular}{l}\includegraphics[width=0.198\linewidth]{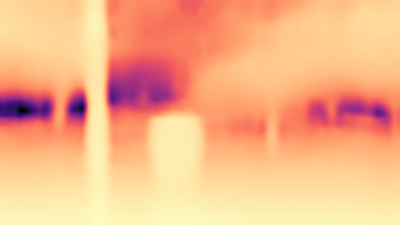}\end{tabular} &
\begin{tabular}{l}\includegraphics[width=0.198\linewidth]{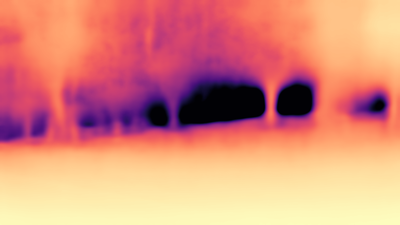}\end{tabular} \\

\rotatebox[origin=c]{90}{Ours} &
\begin{tabular}{l}\includegraphics[width=0.198\linewidth]{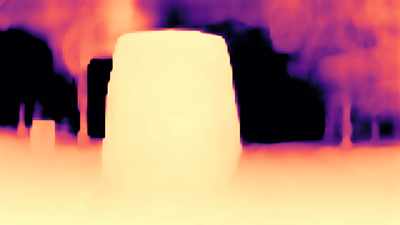}\end{tabular} &
\begin{tabular}{l}\includegraphics[width=0.198\linewidth]{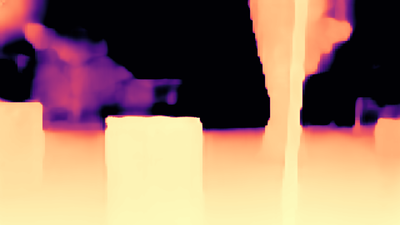}\end{tabular} &
\begin{tabular}{l}\includegraphics[width=0.198\linewidth]{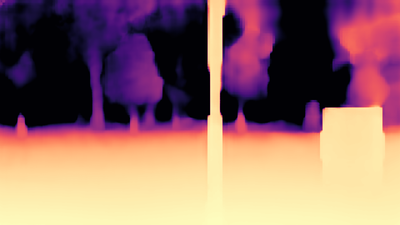}\end{tabular} &
\begin{tabular}{l}\includegraphics[width=0.198\linewidth]{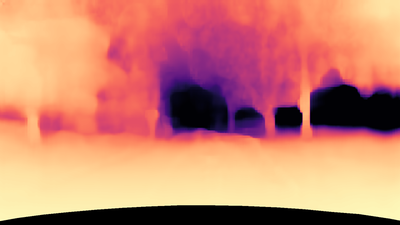}\end{tabular} &
\begin{tabular}{l}\includegraphics[width=0.198\linewidth]{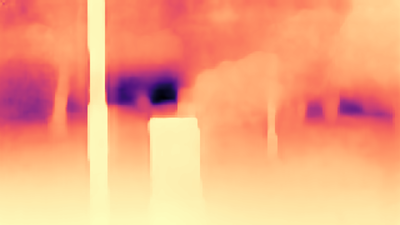}\end{tabular} &
\begin{tabular}{l}\includegraphics[width=0.198\linewidth]{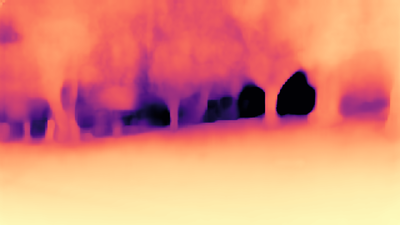}\end{tabular} \\
& \multicolumn{6}{c}{\begin{tabular}{l}\includegraphics[width=0.5\linewidth]{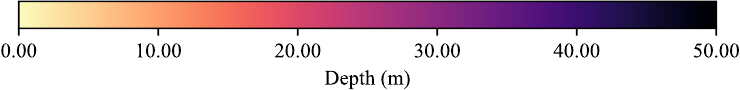}\end{tabular}} \\

\end{tabular}}
\vspace{-0.2cm}
\caption{Qualitative comparison of predicted surrounding depth on NuScenes~\cite{caesar2020nuscenes}. We show a comparison of depth maps from our method to the depth maps of the state-of-the-art approach SurroundDepth~\cite{wei2023surrounddepth}. We observe that our method produces significantly sharper and more accurate depth predictions, particularly in fine details.}
\label{fig:nuscenes_depth_examples}
\end{figure*}
\begin{figure*}[t]
\setlength\tabcolsep{0.5 pt}
\centering
\scalebox{0.8}{
\begin{tabular}{lcccccc}

& Front & F.Left & B.Left & Back & B.Right & F.Right \\
\rotatebox[origin=c]{90}{Input} &
\begin{tabular}{l}\includegraphics[width=0.198\linewidth]{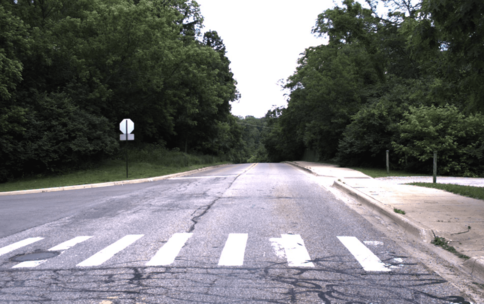}\end{tabular} &
\begin{tabular}{l}\includegraphics[width=0.198\linewidth]{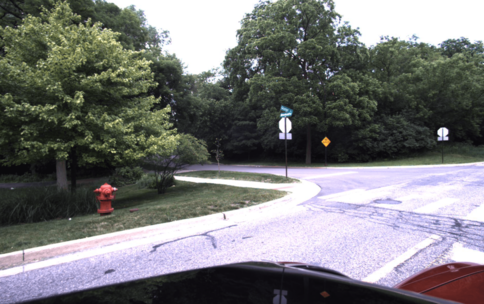}\end{tabular} &
\begin{tabular}{l}\includegraphics[width=0.198\linewidth]{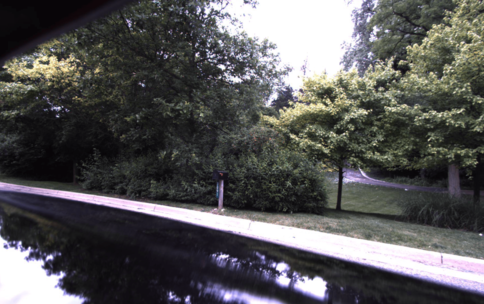}\end{tabular} &
\begin{tabular}{l}\includegraphics[width=0.198\linewidth]{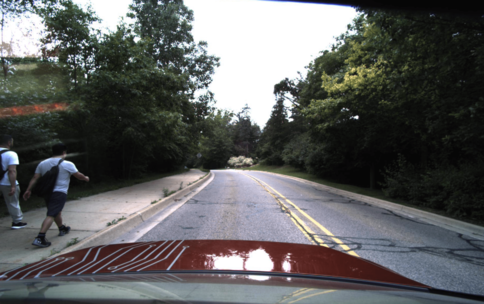}\end{tabular} &
\begin{tabular}{l}\includegraphics[width=0.198\linewidth]{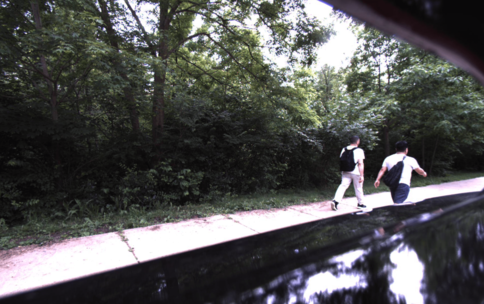}\end{tabular} &
\begin{tabular}{l}\includegraphics[width=0.198\linewidth]{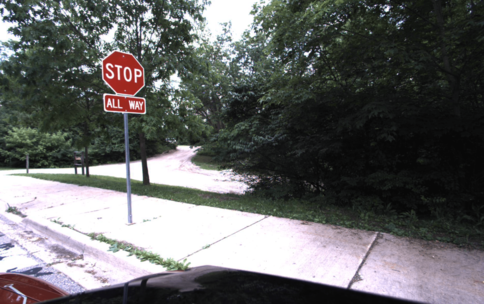}\end{tabular} \\

\rotatebox[origin=c]{90}{SD~\cite{wei2023surrounddepth}} &
\begin{tabular}{l}\includegraphics[width=0.198\linewidth]{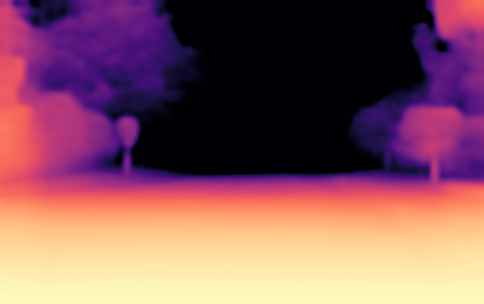}\end{tabular} &
\begin{tabular}{l}\includegraphics[width=0.198\linewidth]{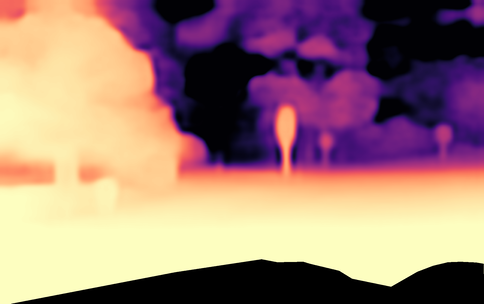}\end{tabular} &
\begin{tabular}{l}\includegraphics[width=0.198\linewidth]{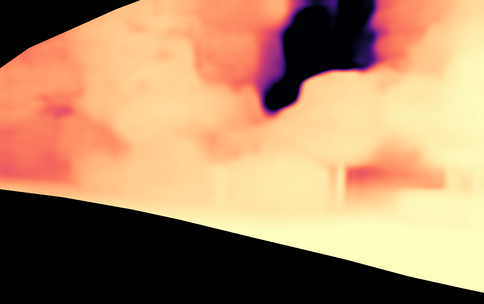}\end{tabular} &
\begin{tabular}{l}\includegraphics[width=0.198\linewidth]{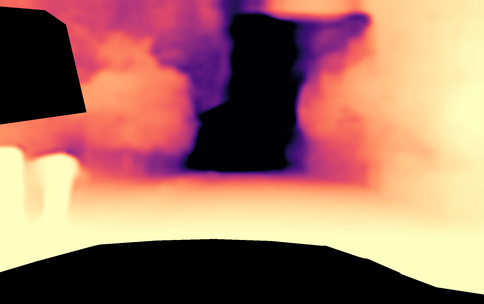}\end{tabular} &
\begin{tabular}{l}\includegraphics[width=0.198\linewidth]{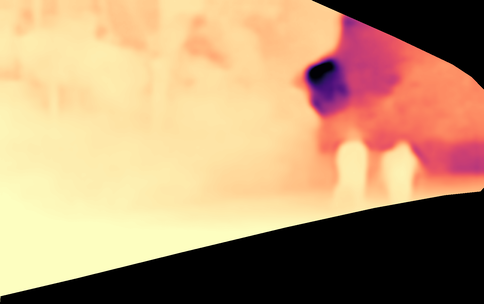}\end{tabular} &
\begin{tabular}{l}\includegraphics[width=0.198\linewidth]{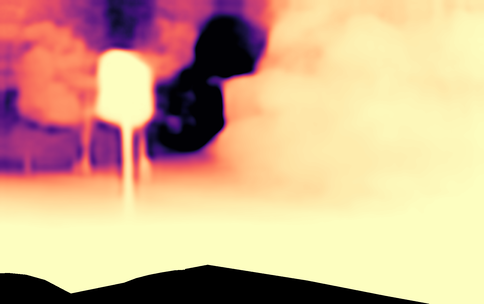}\end{tabular} \\

\rotatebox[origin=c]{90}{Ours} &
\begin{tabular}{l}\includegraphics[width=0.198\linewidth]{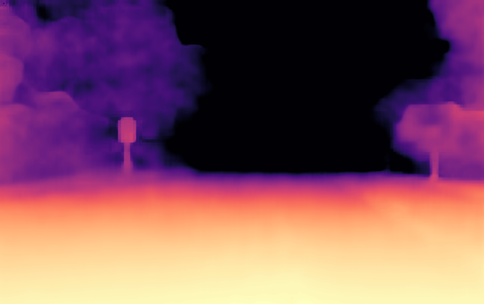}\end{tabular} &
\begin{tabular}{l}\includegraphics[width=0.198\linewidth]{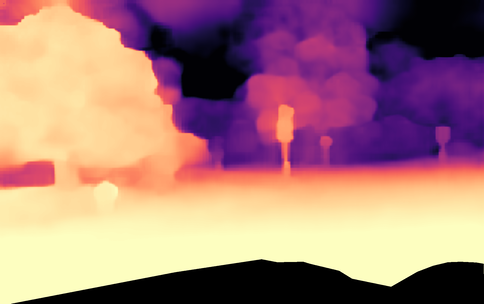}\end{tabular} &
\begin{tabular}{l}\includegraphics[width=0.198\linewidth]{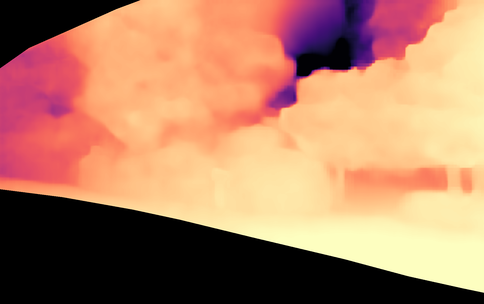}\end{tabular} &
\begin{tabular}{l}\includegraphics[width=0.198\linewidth]{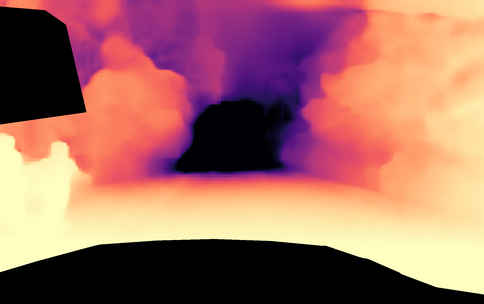}\end{tabular} &
\begin{tabular}{l}\includegraphics[width=0.198\linewidth]{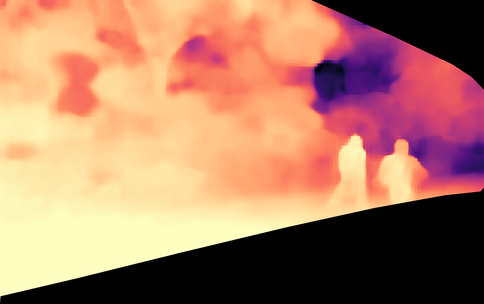}\end{tabular} &
\begin{tabular}{l}\includegraphics[width=0.198\linewidth]{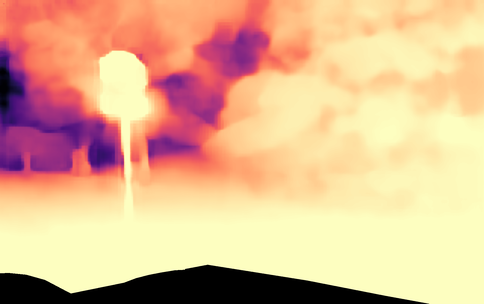}\end{tabular} \\

\midrule

\rotatebox[origin=c]{90}{Input} &
\begin{tabular}{l}\includegraphics[width=0.198\linewidth]{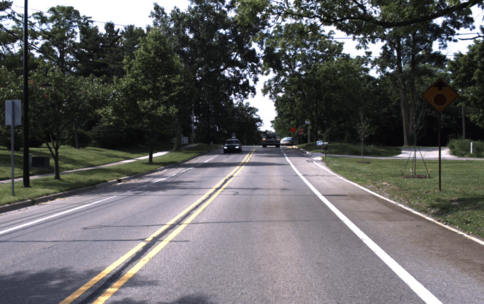}\end{tabular} &
\begin{tabular}{l}\includegraphics[width=0.198\linewidth]{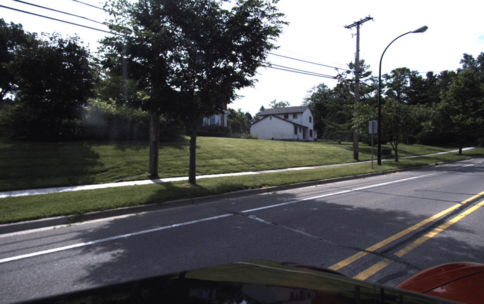}\end{tabular} &
\begin{tabular}{l}\includegraphics[width=0.198\linewidth]{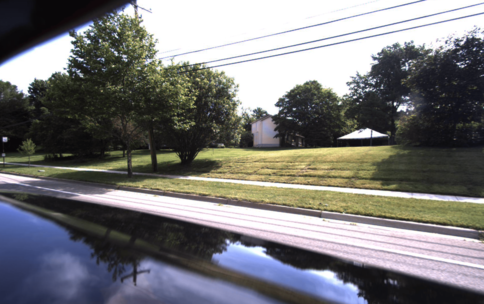}\end{tabular} &
\begin{tabular}{l}\includegraphics[width=0.198\linewidth]{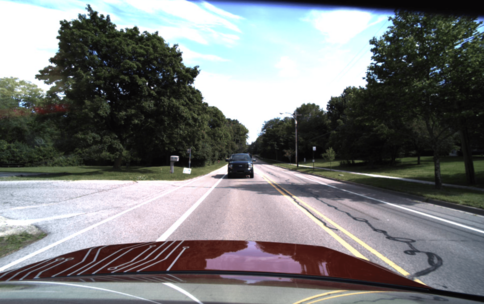}\end{tabular} &
\begin{tabular}{l}\includegraphics[width=0.198\linewidth]{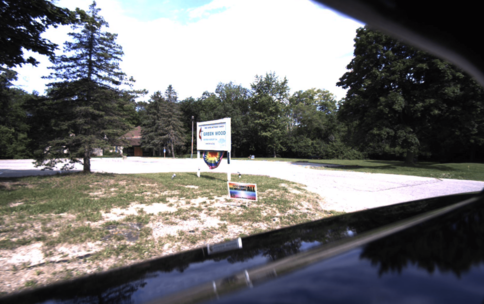}\end{tabular} &
\begin{tabular}{l}\includegraphics[width=0.198\linewidth]{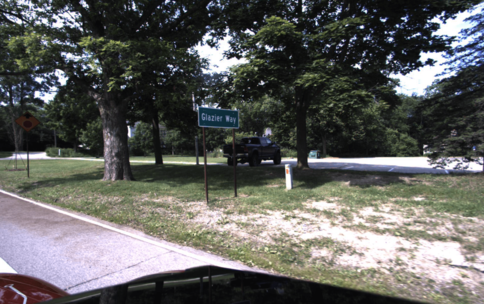}\end{tabular} \\

\rotatebox[origin=c]{90}{SD~\cite{wei2023surrounddepth}} &
\begin{tabular}{l}\includegraphics[width=0.198\linewidth]{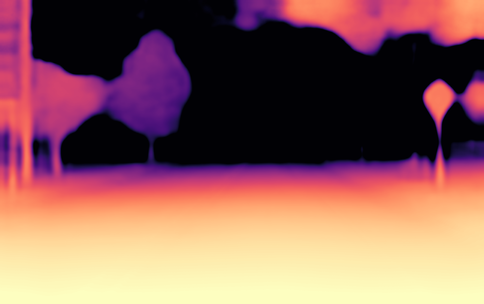}\end{tabular} &
\begin{tabular}{l}\includegraphics[width=0.198\linewidth]{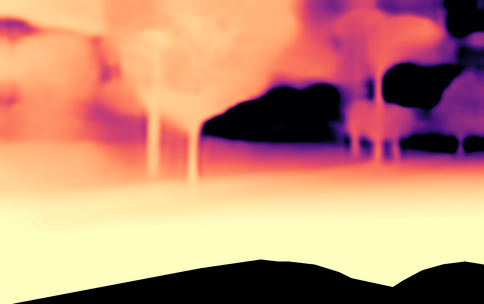}\end{tabular} &
\begin{tabular}{l}\includegraphics[width=0.198\linewidth]{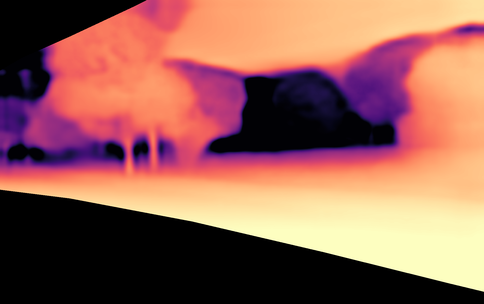}\end{tabular} &
\begin{tabular}{l}\includegraphics[width=0.198\linewidth]{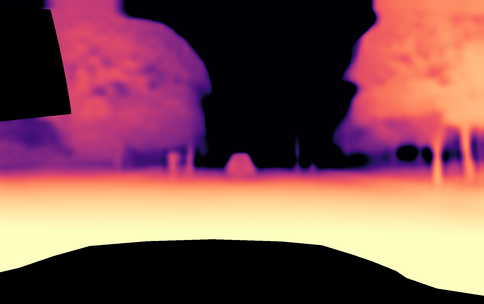}\end{tabular} &
\begin{tabular}{l}\includegraphics[width=0.198\linewidth]{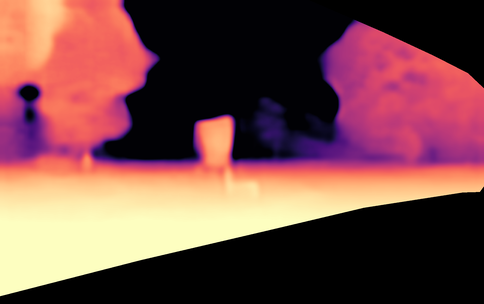}\end{tabular} &
\begin{tabular}{l}\includegraphics[width=0.198\linewidth]{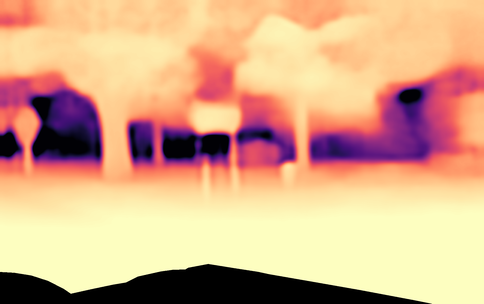}\end{tabular} \\

\rotatebox[origin=c]{90}{Ours} &
\begin{tabular}{l}\includegraphics[width=0.198\linewidth]{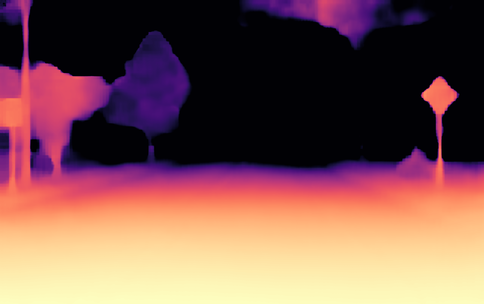}\end{tabular} &
\begin{tabular}{l}\includegraphics[width=0.198\linewidth]{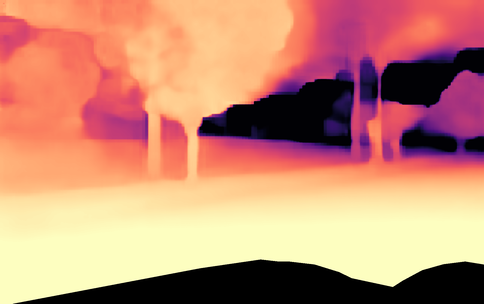}\end{tabular} &
\begin{tabular}{l}\includegraphics[width=0.198\linewidth]{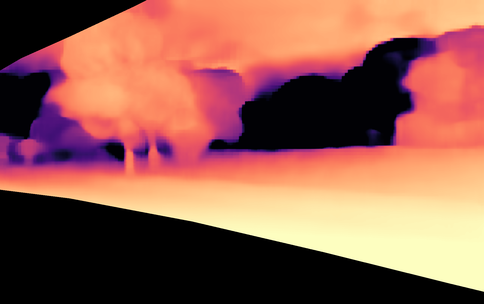}\end{tabular} &
\begin{tabular}{l}\includegraphics[width=0.198\linewidth]{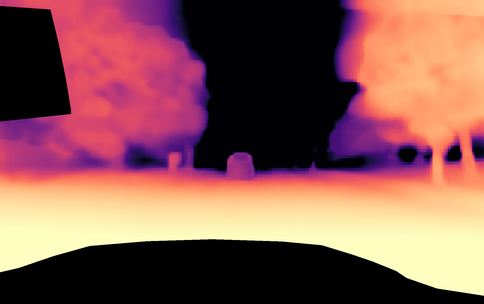}\end{tabular} &
\begin{tabular}{l}\includegraphics[width=0.198\linewidth]{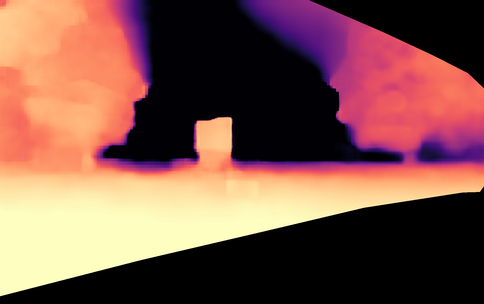}\end{tabular} &
\begin{tabular}{l}\includegraphics[width=0.198\linewidth]{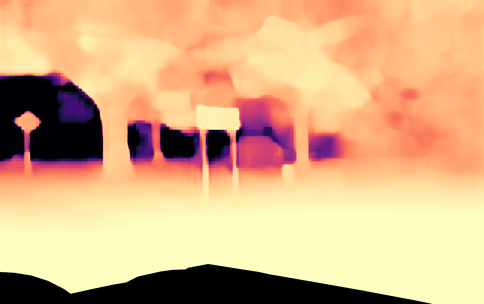}\end{tabular} \\

\midrule

\rotatebox[origin=c]{90}{Input} &
\begin{tabular}{l}\includegraphics[width=0.198\linewidth]{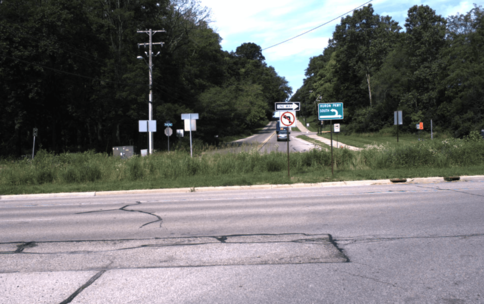}\end{tabular} &
\begin{tabular}{l}\includegraphics[width=0.198\linewidth]{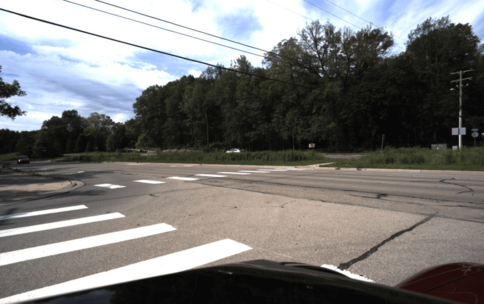}\end{tabular} &
\begin{tabular}{l}\includegraphics[width=0.198\linewidth]{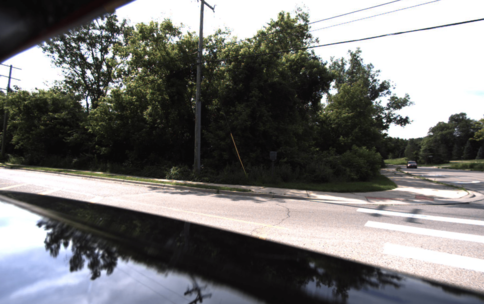}\end{tabular} &
\begin{tabular}{l}\includegraphics[width=0.198\linewidth]{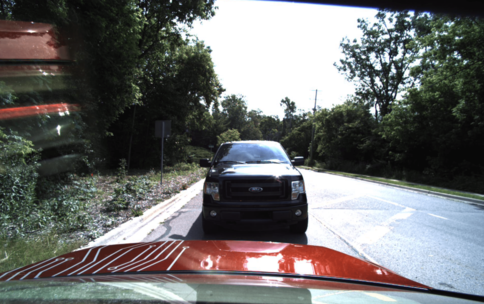}\end{tabular} &
\begin{tabular}{l}\includegraphics[width=0.198\linewidth]{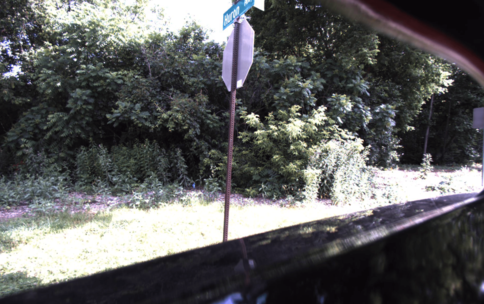}\end{tabular} &
\begin{tabular}{l}\includegraphics[width=0.198\linewidth]{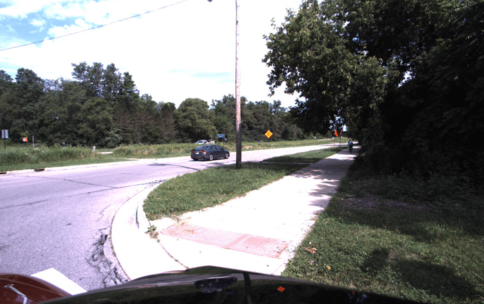}\end{tabular} \\

\rotatebox[origin=c]{90}{SD~\cite{wei2023surrounddepth}} &
\begin{tabular}{l}\includegraphics[width=0.198\linewidth]{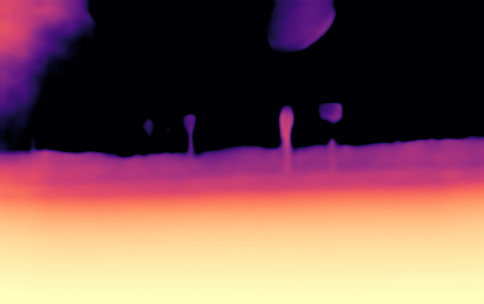}\end{tabular} &
\begin{tabular}{l}\includegraphics[width=0.198\linewidth]{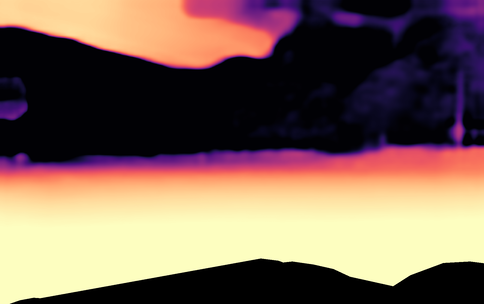}\end{tabular} &
\begin{tabular}{l}\includegraphics[width=0.198\linewidth]{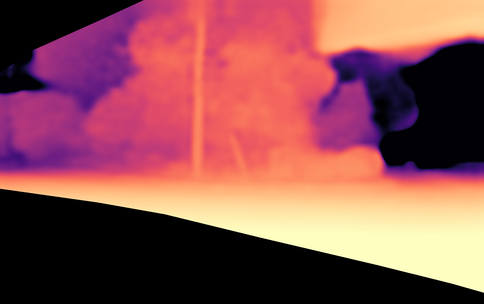}\end{tabular} &
\begin{tabular}{l}\includegraphics[width=0.198\linewidth]{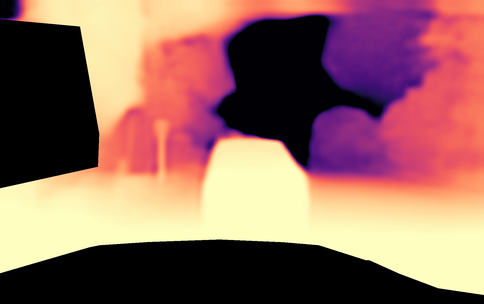}\end{tabular} &
\begin{tabular}{l}\includegraphics[width=0.198\linewidth]{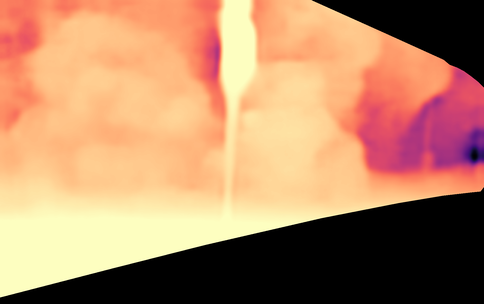}\end{tabular} &
\begin{tabular}{l}\includegraphics[width=0.198\linewidth]{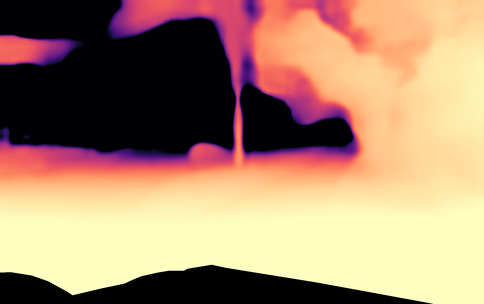}\end{tabular} \\

\rotatebox[origin=c]{90}{Ours} &
\begin{tabular}{l}\includegraphics[width=0.198\linewidth]{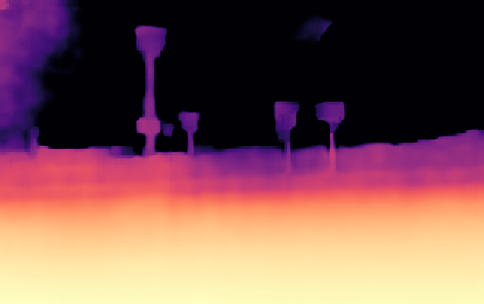}\end{tabular} &
\begin{tabular}{l}\includegraphics[width=0.198\linewidth]{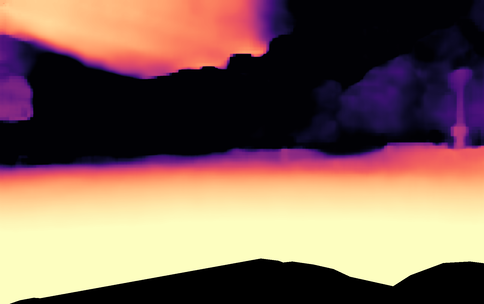}\end{tabular} &
\begin{tabular}{l}\includegraphics[width=0.198\linewidth]{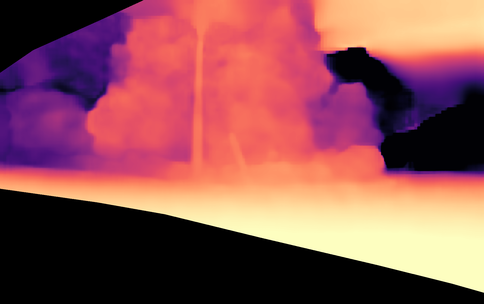}\end{tabular} &
\begin{tabular}{l}\includegraphics[width=0.198\linewidth]{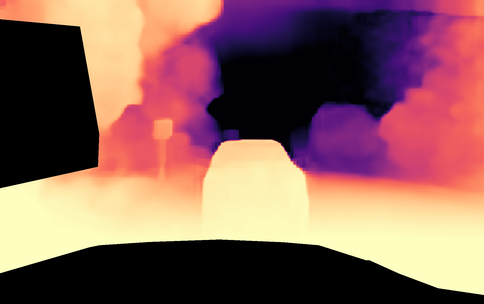}\end{tabular} &
\begin{tabular}{l}\includegraphics[width=0.198\linewidth]{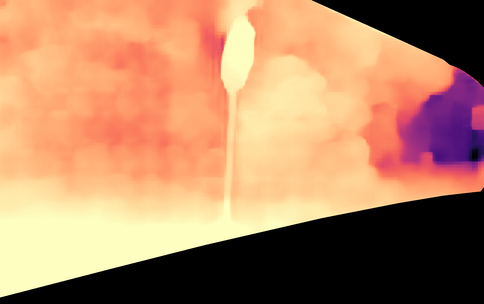}\end{tabular} &
\begin{tabular}{l}\includegraphics[width=0.198\linewidth]{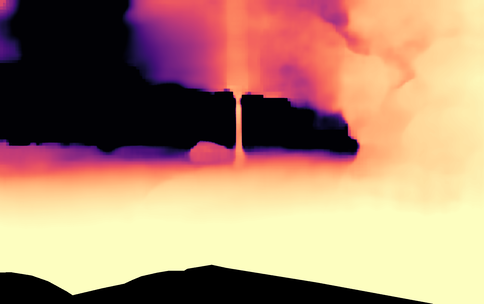}\end{tabular} \\

& & & & & & \\
& \multicolumn{6}{c}{\begin{tabular}{l}\includegraphics[width=0.5\linewidth]{images/sup-img/ddad_error_map/depth_error_colorbar-cropped.pdf}\end{tabular}} \\

\end{tabular}}

\caption{Qualitative comparison of predicted surrounding depth on DDAD~\cite{guizilini2020ddad_packnet}. We show a comparison of depth maps from M$^2$Depth to the depth maps of the state-of-the-art approach SurroundDepth~\cite{wei2023surrounddepth}. We observe that our method produces significantly sharper and more accurate depth predictions, particularly in fine details.}
\label{fig:ddad_depth_examples}
\end{figure*}
We visualize more depth results in Nuscenes~\cite{caesar2020nuscenes} and DDAD~\cite{guizilini2020ddad_packnet} dataset. In ~\cref{fig:nuscenes_depth_examples} and ~\cref{fig:ddad_depth_examples}, our M$^2$Depth consistently exhibits robustness and effectiveness across diverse scenes. Notably, at the object edges, our method produces sharper depth predictions.
\subsection{More Depth Error Results}
\begin{figure*}[t]
\setlength\tabcolsep{0.5 pt}
\centering
\scalebox{0.8}{
\begin{tabular}{lcccccc}

& Front & F.Left & B.Left & Back & B.Right & F.Right \\
\rotatebox[origin=c]{90}{Input} &
\begin{tabular}{l}\includegraphics[width=0.198\linewidth]{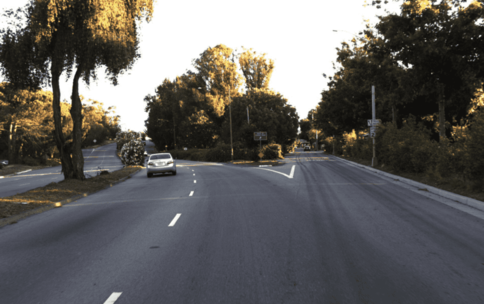}\end{tabular} &
\begin{tabular}{l}\includegraphics[width=0.198\linewidth]{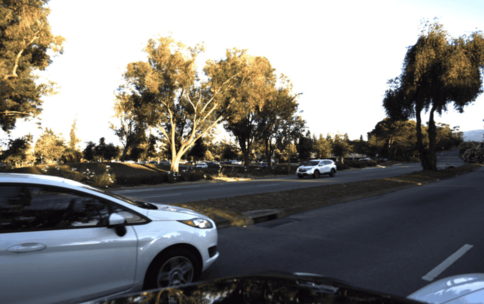}\end{tabular} &
\begin{tabular}{l}\includegraphics[width=0.198\linewidth]{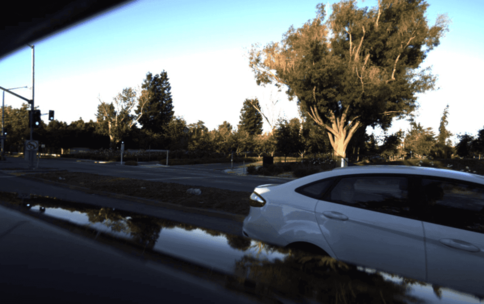}\end{tabular} &
\begin{tabular}{l}\includegraphics[width=0.198\linewidth]{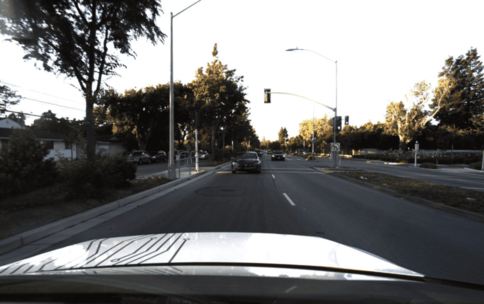}\end{tabular} &
\begin{tabular}{l}\includegraphics[width=0.198\linewidth]{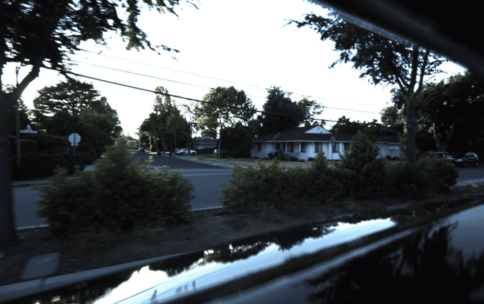}\end{tabular} &
\begin{tabular}{l}\includegraphics[width=0.198\linewidth]{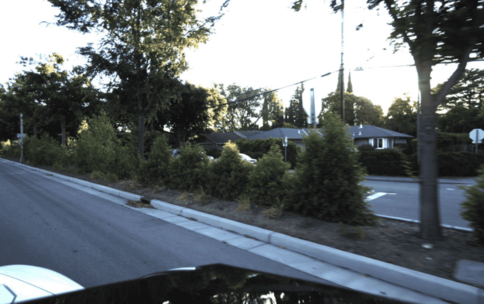}\end{tabular} \\


\multirow{2}{*}[1.2em]{\rotatebox[origin=c]{90}{SurroundDepth~\cite{wei2023surrounddepth}}} &
\begin{tabular}{l}\includegraphics[width=0.198\linewidth]{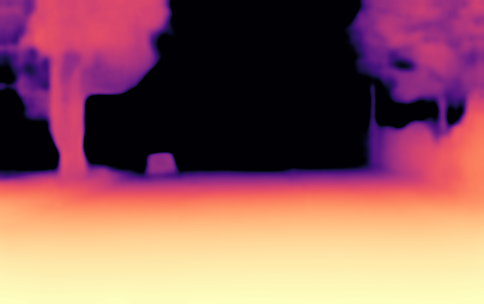}\end{tabular} &
\begin{tabular}{l}\includegraphics[width=0.198\linewidth]{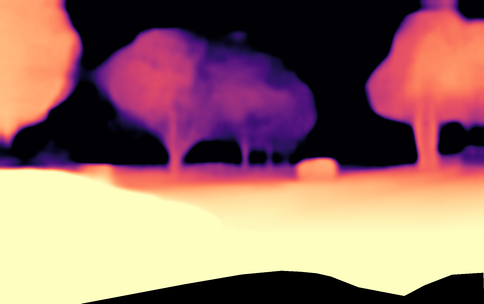}\end{tabular} &
\begin{tabular}{l}\includegraphics[width=0.198\linewidth]{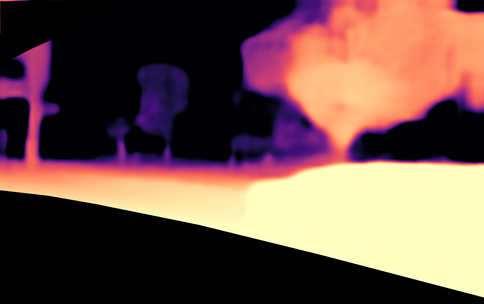}\end{tabular} &
\begin{tabular}{l}\includegraphics[width=0.198\linewidth]{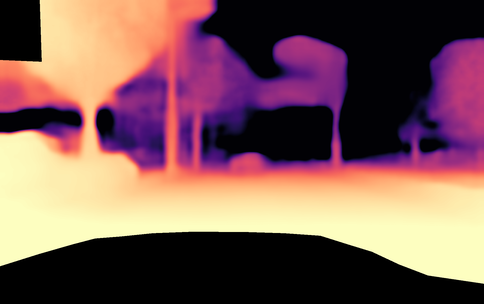}\end{tabular} &
\begin{tabular}{l}\includegraphics[width=0.198\linewidth]{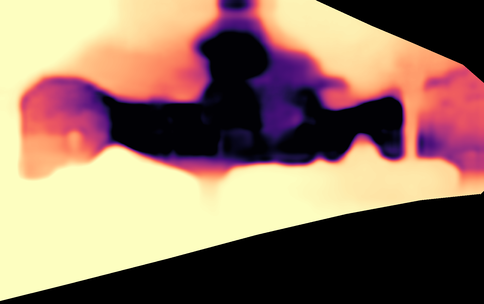}\end{tabular} &
\begin{tabular}{l}\includegraphics[width=0.198\linewidth]{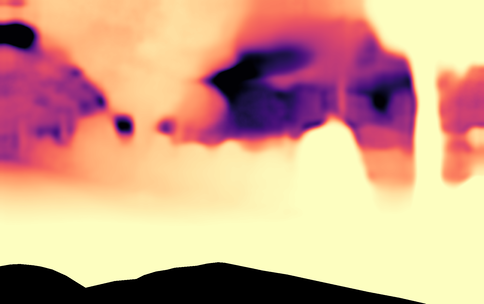}\end{tabular} \\
&
\begin{tabular}{l}\includegraphics[width=0.198\linewidth]{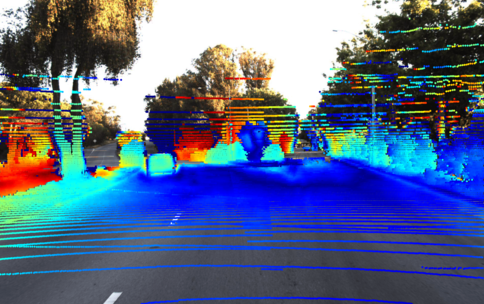}\end{tabular} &
\begin{tabular}{l}\includegraphics[width=0.198\linewidth]{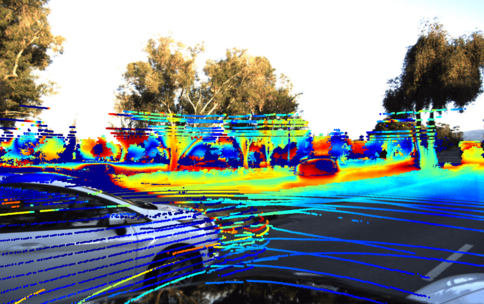}\end{tabular} &
\begin{tabular}{l}\includegraphics[width=0.198\linewidth]{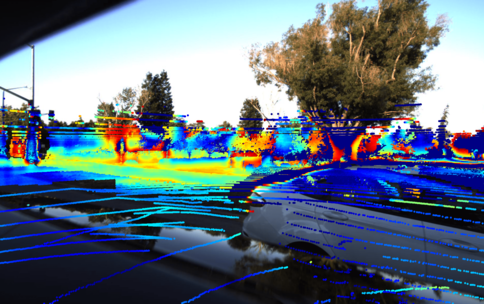}\end{tabular} &
\begin{tabular}{l}\includegraphics[width=0.198\linewidth]{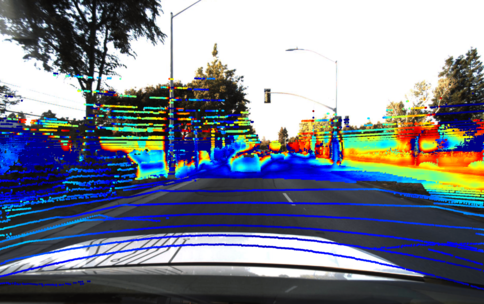}\end{tabular} &
\begin{tabular}{l}\includegraphics[width=0.198\linewidth]{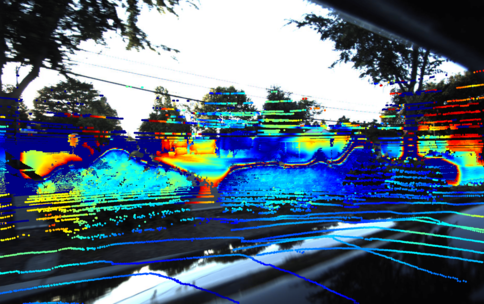}\end{tabular} &
\begin{tabular}{l}\includegraphics[width=0.198\linewidth]{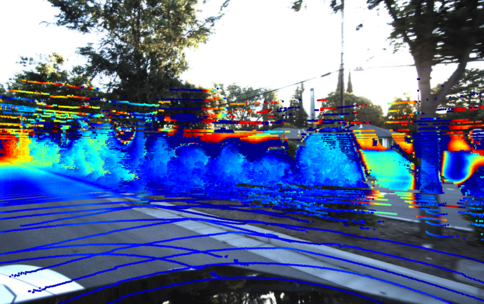}\end{tabular} \\


\multirow{2}{*}[-1.5em]{\rotatebox[origin=c]{90}{Ours}} &
\begin{tabular}{l}\includegraphics[width=0.198\linewidth]{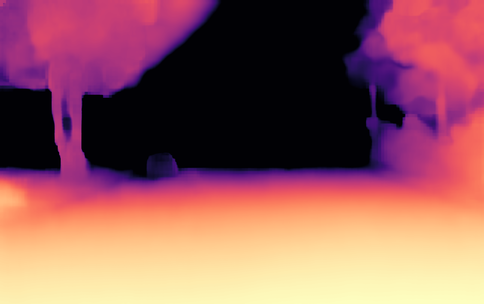}\end{tabular} &
\begin{tabular}{l}\includegraphics[width=0.198\linewidth]{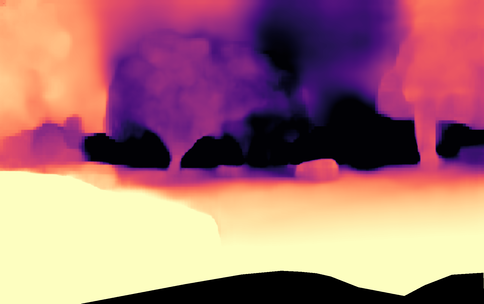}\end{tabular} &
\begin{tabular}{l}\includegraphics[width=0.198\linewidth]{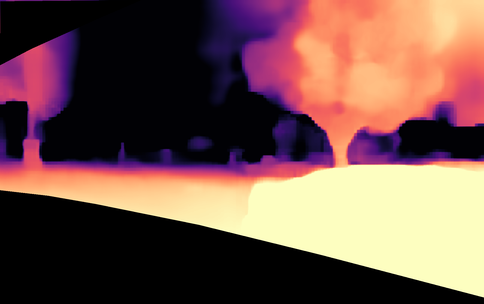}\end{tabular} &
\begin{tabular}{l}\includegraphics[width=0.198\linewidth]{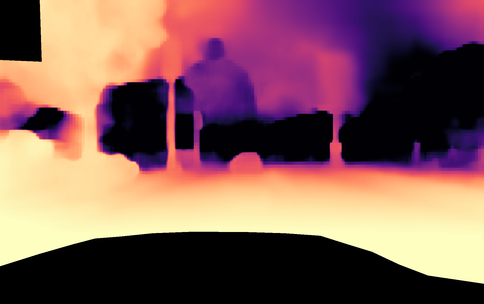}\end{tabular} &
\begin{tabular}{l}\includegraphics[width=0.198\linewidth]{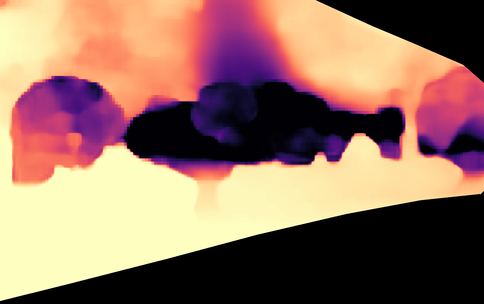}\end{tabular} &
\begin{tabular}{l}\includegraphics[width=0.198\linewidth]{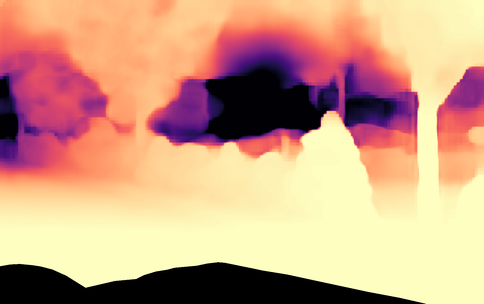}\end{tabular} \\
&
\begin{tabular}{l}\includegraphics[width=0.198\linewidth]{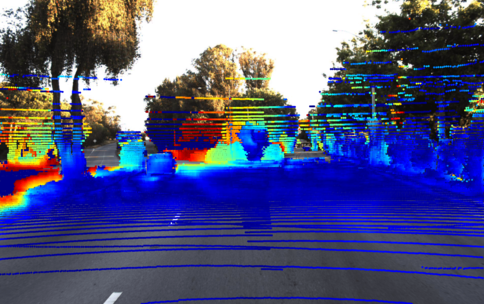}\end{tabular} &
\begin{tabular}{l}\includegraphics[width=0.198\linewidth]{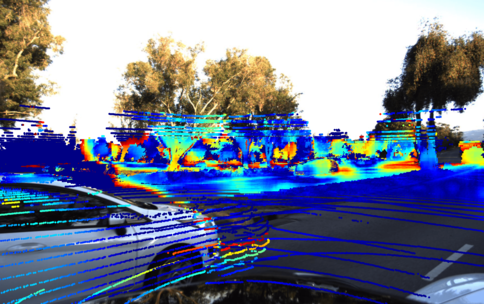}\end{tabular} &
\begin{tabular}{l}\includegraphics[width=0.198\linewidth]{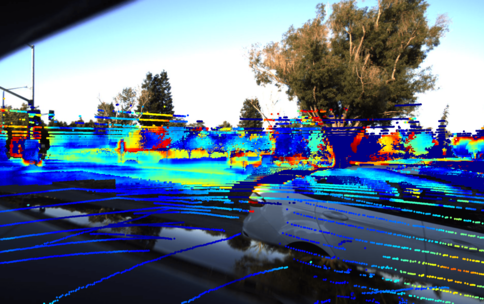}\end{tabular} &
\begin{tabular}{l}\includegraphics[width=0.198\linewidth]{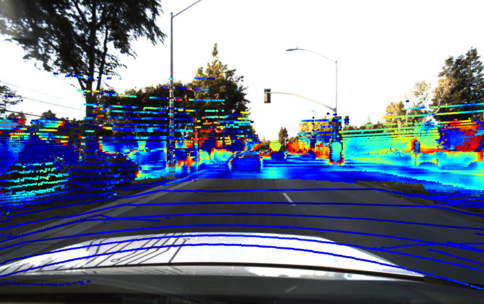}\end{tabular} &
\begin{tabular}{l}\includegraphics[width=0.198\linewidth]{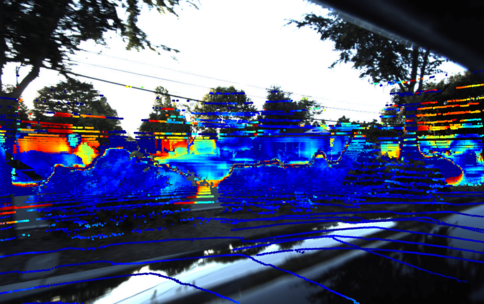}\end{tabular} &
\begin{tabular}{l}\includegraphics[width=0.198\linewidth]{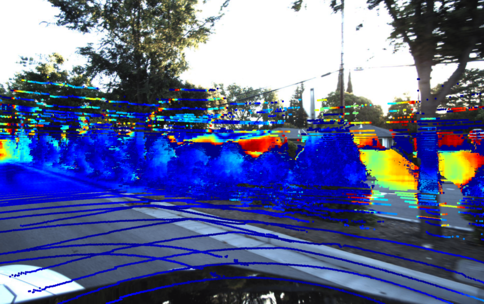}\end{tabular} \\

\midrule

\rotatebox[origin=c]{90}{Input} &
\begin{tabular}{l}\includegraphics[width=0.198\linewidth]{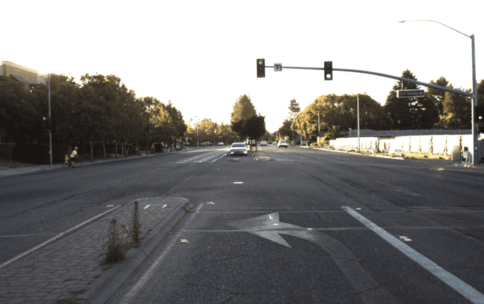}\end{tabular} &
\begin{tabular}{l}\includegraphics[width=0.198\linewidth]{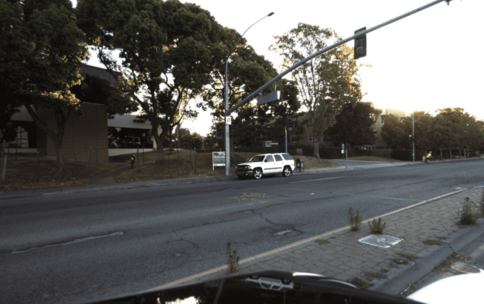}\end{tabular} &
\begin{tabular}{l}\includegraphics[width=0.198\linewidth]{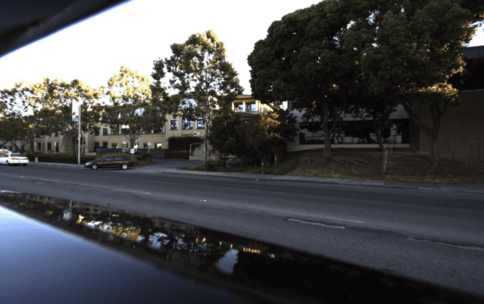}\end{tabular} &
\begin{tabular}{l}\includegraphics[width=0.198\linewidth]{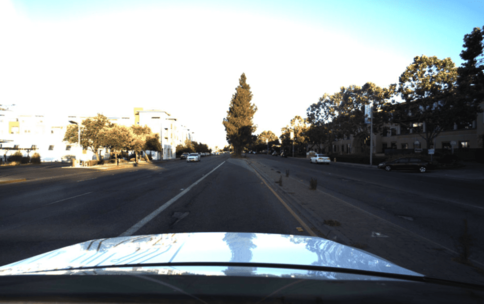}\end{tabular} &
\begin{tabular}{l}\includegraphics[width=0.198\linewidth]{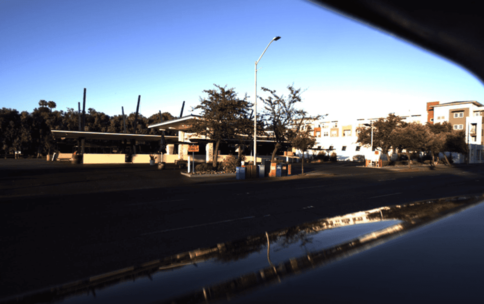}\end{tabular} &
\begin{tabular}{l}\includegraphics[width=0.198\linewidth]{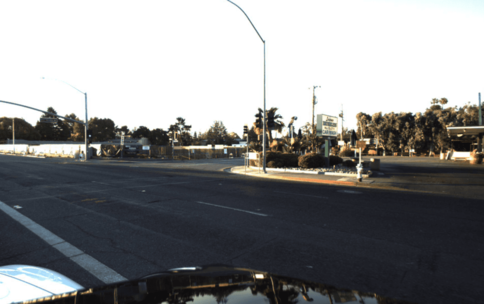}\end{tabular} \\


\multirow{2}{*}[1.2em]{\rotatebox[origin=c]{90}{SurroundDepth~\cite{wei2023surrounddepth}}} &
\begin{tabular}{l}\includegraphics[width=0.198\linewidth]{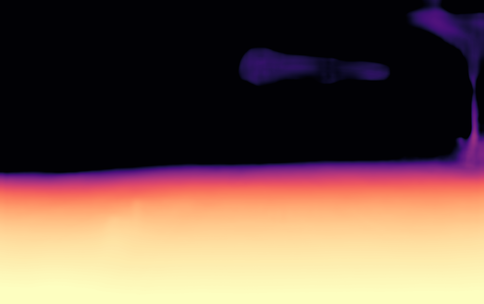}\end{tabular} &
\begin{tabular}{l}\includegraphics[width=0.198\linewidth]{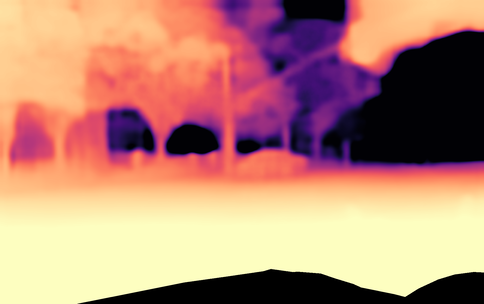}\end{tabular} &
\begin{tabular}{l}\includegraphics[width=0.198\linewidth]{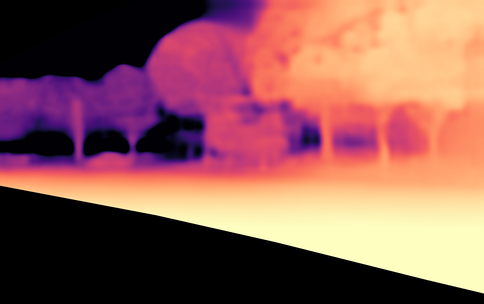}\end{tabular} &
\begin{tabular}{l}\includegraphics[width=0.198\linewidth]{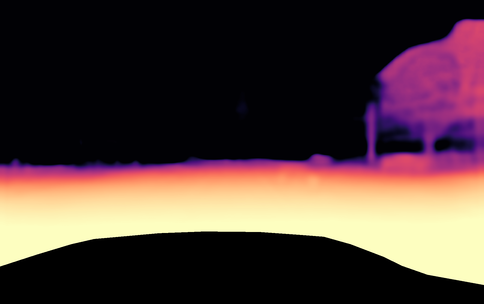}\end{tabular} &
\begin{tabular}{l}\includegraphics[width=0.198\linewidth]{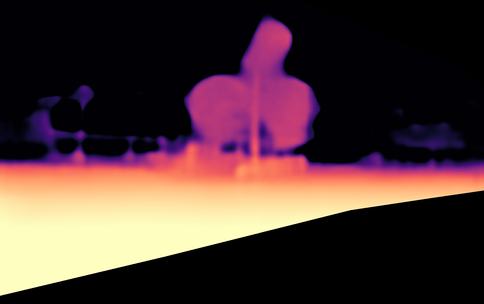}\end{tabular} &
\begin{tabular}{l}\includegraphics[width=0.198\linewidth]{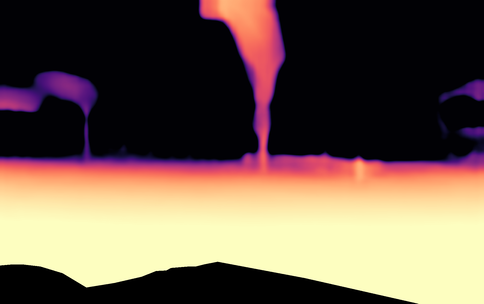}\end{tabular} \\
&
\begin{tabular}{l}\includegraphics[width=0.198\linewidth]{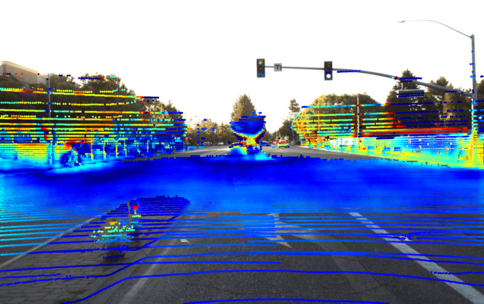}\end{tabular} &
\begin{tabular}{l}\includegraphics[width=0.198\linewidth]{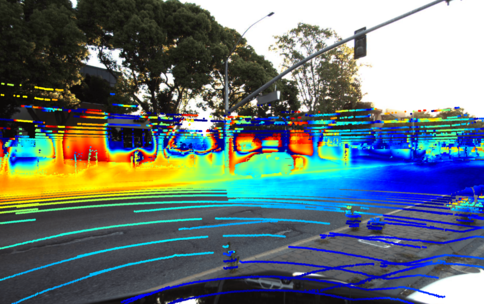}\end{tabular} &
\begin{tabular}{l}\includegraphics[width=0.198\linewidth]{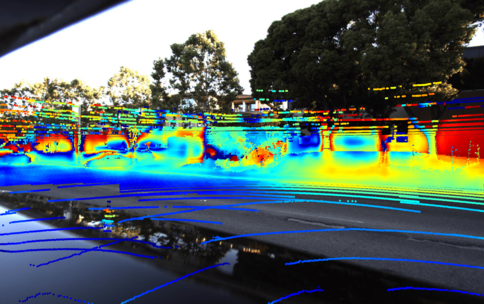}\end{tabular} &
\begin{tabular}{l}\includegraphics[width=0.198\linewidth]{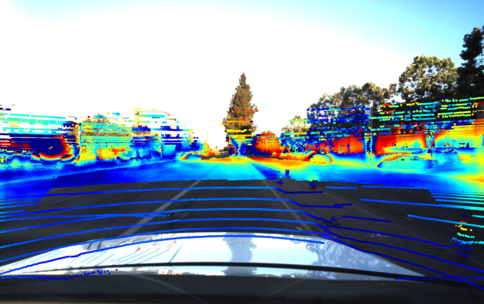}\end{tabular} &
\begin{tabular}{l}\includegraphics[width=0.198\linewidth]{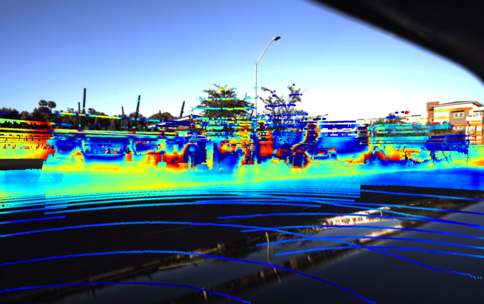}\end{tabular} &
\begin{tabular}{l}\includegraphics[width=0.198\linewidth]{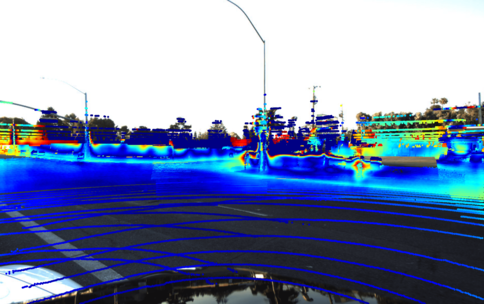}\end{tabular} \\


\multirow{2}{*}[-1.5em]{\rotatebox[origin=c]{90}{Ours}} &
\begin{tabular}{l}\includegraphics[width=0.198\linewidth]{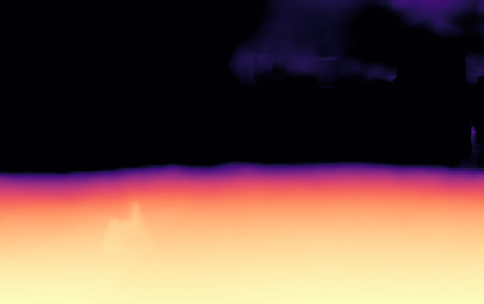}\end{tabular} &
\begin{tabular}{l}\includegraphics[width=0.198\linewidth]{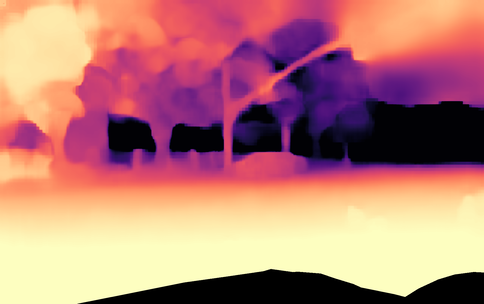}\end{tabular} &
\begin{tabular}{l}\includegraphics[width=0.198\linewidth]{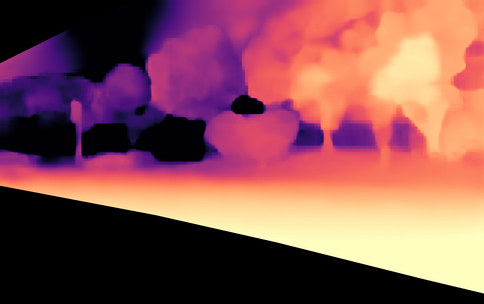}\end{tabular} &
\begin{tabular}{l}\includegraphics[width=0.198\linewidth]{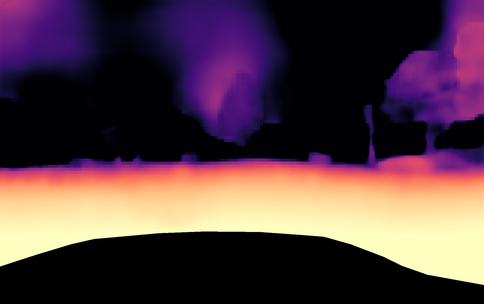}\end{tabular} &
\begin{tabular}{l}\includegraphics[width=0.198\linewidth]{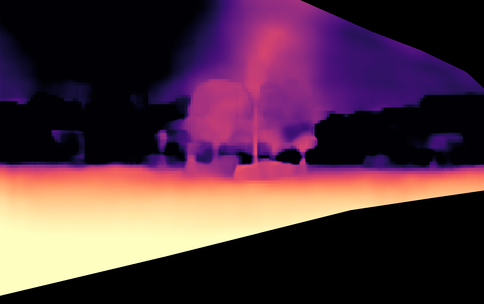}\end{tabular} &
\begin{tabular}{l}\includegraphics[width=0.198\linewidth]{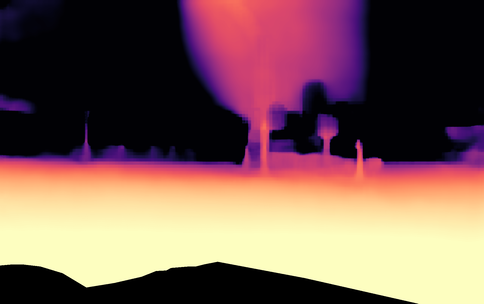}\end{tabular} \\
&
\begin{tabular}{l}\includegraphics[width=0.198\linewidth]{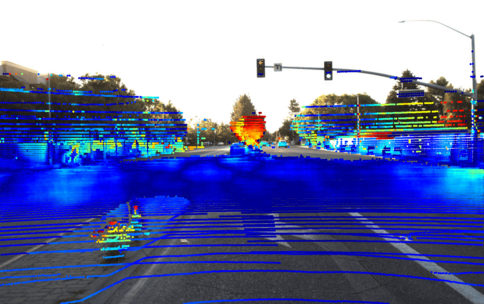}\end{tabular} &
\begin{tabular}{l}\includegraphics[width=0.198\linewidth]{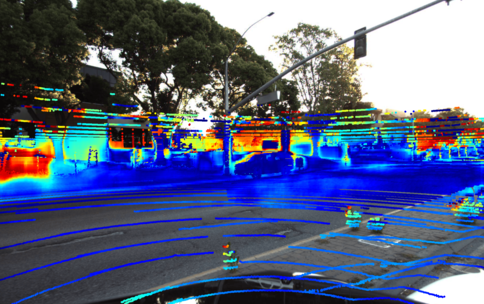}\end{tabular} &
\begin{tabular}{l}\includegraphics[width=0.198\linewidth]{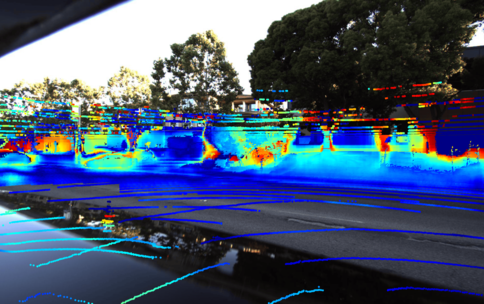}\end{tabular} &
\begin{tabular}{l}\includegraphics[width=0.198\linewidth]{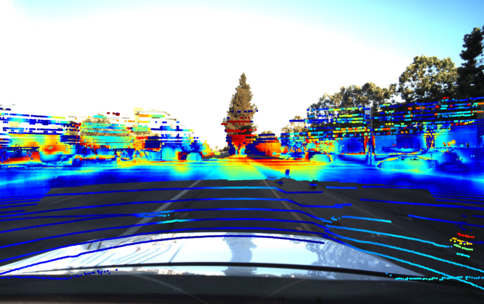}\end{tabular} &
\begin{tabular}{l}\includegraphics[width=0.198\linewidth]{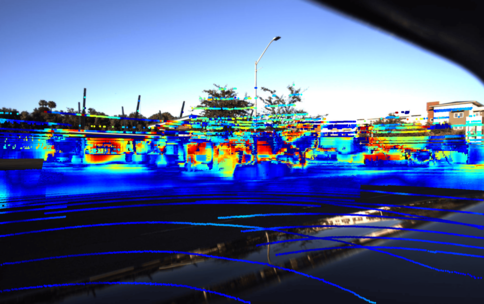}\end{tabular} &
\begin{tabular}{l}\includegraphics[width=0.198\linewidth]{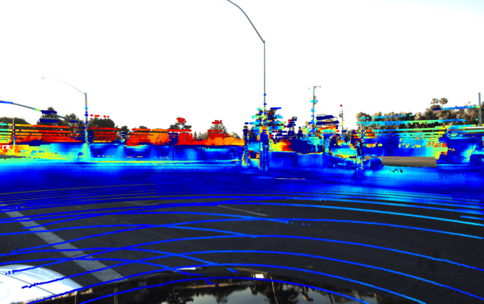}\end{tabular} \\
&
\multicolumn{3}{c}{\begin{tabular}{l}\includegraphics[width=0.5\linewidth]{images/sup-img/ddad_error_map/depth_error_colorbar-cropped.pdf}\end{tabular}} &
\multicolumn{3}{c}{\begin{tabular}{l}\includegraphics[width=0.5\linewidth]{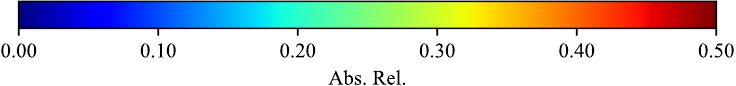}\end{tabular}} \\
\end{tabular}}
\caption{Qualitative comparison of predicted surrounding depth on DDAD dataset~\cite{guizilini2020ddad_packnet}. 
Given the input surrounding images (the top row), we show the visualized depth maps and depth errors of SurroundDepth~\cite{wei2023surrounddepth} and M$^2$Depth. Our method is able to produce more accurate depth with less error and sharper depth edge across multiple cameras.
}
\label{fig:ddad_error_map_sup}
\end{figure*}
In ~\cref{fig:ddad_error_map_sup}, we qualitatively compare our method with existing works in terms of scale-aware depth estimation in DDAD. It can be observed that our method achieves better results at the overlapping between adjacent views.

\par\vfill\par
\clearpage  

%
%
\bibliographystyle{splncs04}
\bibliography{main}
\end{document}